%% file: main.tex
\documentclass[11pt, a4paper]{lumia}

\usepackage[sort&compress]{natbib}
\input{notations}

\usepackage{graphicx}           
\usepackage{tikz}               
\usepackage[edges]{forest}      

\usepackage{url}                
\usepackage{xurl}               

\usepackage{array}              
\usepackage{longtable}          
\usepackage{multirow}           
\usepackage{makecell}           
\usepackage{ragged2e}           

\usepackage{mathtools}          
\usepackage{nicefrac}           

\usepackage{algorithm}          
\usepackage{algorithmicx}       
\usepackage{algpseudocode}      
\usepackage{listings}           

\usepackage{subcaption}         
\usepackage{wrapfig}            
\usepackage[export]{adjustbox}  

\usepackage{changes}            
\usepackage{xspace}             
\usepackage[normalem]{ulem}     
\usepackage{CJKutf8}            

\usepackage[tikz]{bclogo}       
\usepackage[framemethod=tikz]{mdframed} 

\usepackage{lipsum}             
\usepackage{tocloft}            
\usepackage{afterpage}          
\usepackage{bbding}             
\usepackage{epigraph}           
\usepackage{minitoc}            
\usepackage{multicol}           
\usepackage{textgreek}          

\newcolumntype{P}[1]{>{\RaggedRight\arraybackslash}p{#1}}

\definecolor{uclablue}{RGB}{39, 116, 174}
\definecolor{bigaired}{RGB}{156, 0, 0}
\definecolor{myblue}{HTML}{598BE7}
\definecolor{mildblue}{RGB}{31,119,180}
\definecolor{sectionblue}{RGB}{70, 130, 180}
\definecolor{methodblue}{RGB}{0, 150, 136}
\definecolor{bgblue}{RGB}{245,243,253}
\definecolor{ttblue}{RGB}{91,194,224}
\definecolor{mygreen}{rgb}{0.64, 0.56, 0.88}
\definecolor{myyellow}{rgb}{0.68, 0.6, 0.1}
\definecolor{fancygreen}{rgb}{0.33, 0.68, 0.20}
\definecolor{salmon}{rgb}{0.94, 0.52, 0.49}
\definecolor{tablegreen}{rgb}{0.82, 0.94, 0.75}
\definecolor{tableblue}{rgb}{0.81, 0.90, 0.94}
\definecolor{tablered}{rgb}{0.97, 0.85, 0.85}
\definecolor{tableorange}{rgb}{0.96, 0.85, 0.81}
\definecolor{myorange}{rgb}{1.0, 0.49, 0.0}
\definecolor{tlgreen}{rgb}{0.33, 0.68, 0.20}
\definecolor{darkgreen}{RGB}{0,100,0}
\definecolor{darkred}{RGB}{200, 0, 0}
\definecolor{customyellow}{HTML}{FFFACD}
\definecolor{refinegreen}{RGB}{0, 128, 75}
\definecolor{scoregreen}{RGB}{34, 139, 34}
\definecolor{hidden-blue}{RGB}{194,232,247}
\definecolor{hidden-black}{RGB}{20,68,106}
\definecolor{yes}{HTML}{C6EFCE}
\definecolor{no}{HTML}{FFC7CE}
\definecolor{partial}{HTML}{FFEB9C}
\definecolor{external}{HTML}{D9E1F2}
\definecolor{hdr}{HTML}{F2F2F2}
\definecolor{GRPOrow}{gray}{0.96}
\definecolor{FlowRLrow}{RGB}{225,236,255}
\definecolor{FlowBlue}{RGB}{80,120,210}
\definecolor{GRPOGray}{gray}{0.35}

\setlist[itemize]{leftmargin=20pt, noitemsep, topsep=0pt}

\NewDocumentCommand{\kaiyan}{mO{}}{\textcolor{purple}{\textsuperscript{\textit{kaiyan}}\textsf{\textbf{\small[#1]}}}}
\NewDocumentCommand{\yuxin}{mO{}}{\textcolor{cyan}{\textsuperscript{\textit{yuxin}}\textsf{\textbf{\small[#1]}}}}
\NewDocumentCommand{\bx}{mO{}}{\textcolor{green}{\textsuperscript{\textit{bx}}\textsf{\textbf{\small[#1]}}}}
\NewDocumentCommand{\at}{mO{}}{\textcolor{red}{\textsuperscript{\textit{AT}}\textsf{\textbf{\small[#1]}}}}
\NewDocumentCommand{\re}{mO{}}{\textcolor{blue}{\textsuperscript{\textit{RE}}\textsf{\textbf{\small[#1]}}}}
\NewDocumentCommand{\ybsun}{mO{}}{\textcolor{magenta}{\textsuperscript{\textit{youbang}}\textsf{\textbf{\small[#1]}}}}
\NewDocumentCommand{\runze}{mO{}}{\textcolor{orange}{\textsuperscript{\textit{runze}}\textsf{\textbf{\small[#1]}}}}
\NewDocumentCommand{\add}{mO{}}{\textcolor{darkgreen}{\textsuperscript{\textit{Maybe Consider Discuss}}\textsf{\textbf{[#1]}}}}

\newcommand{\cmark}{\textcolor{darkgreen}{\boldmath$\checkmark$}}
\newcommand{\xmark}{\textcolor{darkred}{\boldmath$\times$}}

\newenvironment{itemize*}%
 {\leftmargini=10pt\begin{itemize}%
  \setlength{\itemsep}{0pt}%
  \setlength{\parskip}{0pt}%
  }%
 {\end{itemize}}

\newenvironment{enumerate*}%
 {\begin{enumerate}%
  \setlength{\itemsep}{0pt}%
  \setlength{\parskip}{0pt}}%
 {\end{enumerate}}

\newcommand{\cellstatus}[1]{%
  \begingroup
  \StrTrim{#1}[\statusval]%
  \IfStrEq{\statusval}{Yes}{\cellcolor{yes}\cmark}{}%
  \IfStrEq{\statusval}{No}{\cellcolor{no}\xmark}{}%
  \IfBeginWith{\statusval}{Yes (}{\cellcolor{yes}\cmark~\textit{\statusval\unskip}}{}%
  \IfStrEq{\statusval}{Partial}{\cellcolor{partial}\textbf{Partial}}{}%
  \IfStrEq{\statusval}{External}{\cellcolor{external}\textbf{External}}{}%
  \endgroup
}

\newtcolorbox{myboxi}[1][]{
  breakable,
  title=#1,
  colback=red!5,
  colbacktitle=red!5,
  coltitle=black,
  fonttitle=\bfseries,
  bottomrule=0pt,
  toprule=0pt,
  leftrule=2pt,
  rightrule=2pt,
  titlerule=0pt,
  arc=0pt,
  outer arc=0pt,
  colframe=red,
}

\newtcolorbox{myboxnote}[1][]{
  breakable,
  title=#1,
  colback=orange!0,
  colbacktitle=orange!0,
  coltitle=black,
  fonttitle=\bfseries,
  bottomrule=0pt,
  toprule=0pt,
  leftrule=2pt,
  rightrule=2pt,
  titlerule=0pt,
  arc=0pt,
  outer arc=0pt,
  colframe=orange,
}

\newtcolorbox{myboxii}[1][]{
  breakable,
  freelance,
  title=#1,
  colback=white,
  colbacktitle=white,
  coltitle=black,
  fonttitle=\bfseries,
  bottomrule=0pt,
  boxrule=0pt,
  colframe=white,
  overlay unbroken and first={
  \draw[red!75!black,line width=3pt]
    ([xshift=5pt]frame.north west) -- 
    (frame.north west) -- 
    (frame.south west);
  \draw[red!75!black,line width=3pt]
    ([xshift=-5pt]frame.north east) -- 
    (frame.north east) -- 
    (frame.south east);
  },
  overlay unbroken app={
  \draw[red!75!black,line width=3pt,line cap=rect]
    (frame.south west) -- 
    ([xshift=5pt]frame.south west);
  \draw[red!75!black,line width=3pt,line cap=rect]
    (frame.south east) -- 
    ([xshift=-5pt]frame.south east);
  },
  overlay middle and last={
  \draw[red!75!black,line width=3pt]
    (frame.north west) -- 
    (frame.south west);
  \draw[red!75!black,line width=3pt]
    (frame.north east) -- 
    (frame.south east);
  },
  overlay last app={
  \draw[red!75!black,line width=3pt,line cap=rect]
    (frame.south west) --
    ([xshift=5pt]frame.south west);
  \draw[red!75!black,line width=3pt,line cap=rect]
    (frame.south east) --
    ([xshift=-5pt]frame.south east);
  },
}

\tcbset{
  takeawaysbox/.style={
    title=Takeaways,
    colback=lightblue!80,
    colframe=black,
    fonttitle=\bfseries\small,
    coltitle=white,
    colbacktitle=black,
    enhanced,
    attach boxed title to top left={xshift=2.5mm,yshift=-2.5mm},
    boxed title style={rounded corners, size=small, colframe=black, colback=black},
    width=\linewidth,
    arc=3.5mm
  }
}

\mdfdefinestyle{mystyle}{%
  rightline=true,
  innerleftmargin=10,
  innerrightmargin=10,
  outerlinewidth=3pt,
  topline=false,
  rightline=true,
  bottomline=false,
  skipabove=\topsep,
  skipbelow=\topsep
}

\tikzset{%
    every node/.style={font=\tiny},
    parent/.style =          {align=center,text width=2cm,rounded corners=3pt, line width=0.3mm, fill=gray!10,draw=gray!80},
    child/.style =           {align=center,text width=2.0cm,rounded corners=3pt, fill=blue!10,draw=blue!80,line width=0.3mm},
    grandchild/.style =      {align=center,text width=2cm,rounded corners=3pt},
    greatgrandchild/.style = {align=center,text width=1.5cm,rounded corners=3pt},
    greatgrandchild2/.style = {align=center,text width=1.5cm,rounded corners=3pt},    
    referenceblock/.style =  {align=center,text width=1.5cm,rounded corners=2pt},
    pretrain/.style =           {align=center,text width=2.0cm,rounded corners=3pt, fill=blue!10,draw=blue!80,line width=0.3mm},   
    pretrain_work/.style =           {align=center, text width=8.5cm,rounded corners=3pt, fill=blue!10,draw=blue!0,line width=0.3mm},  
    template/.style =           {align=center,text width=2.0cm,rounded corners=3pt, fill=red!10,draw=red!80,line width=0.3mm},   
    template_work/.style =           {align=center,text width=8.5cm,rounded corners=3pt, fill=red!10,draw=red!0,line width=0.3mm},    
    answer/.style =           {align=center,text width=2.0cm,rounded corners=3pt, fill= cyan!10,draw= cyan!80,line width=0.3mm},   
    answer_work/.style =           {align=center,text width=8.5cm,rounded corners=3pt, fill= cyan!10,draw= cyan!0,line width=0.3mm},      
    multiple/.style =           {align=center,text width=2.0cm,rounded corners=3pt, fill= orange!10,draw= orange!80,line width=0.3mm},   
    multiple_work/.style =           {align=center,text width=8.5cm,rounded corners=3pt, fill= orange!10,draw= orange!0,line width=0.3mm},        
    tuning/.style =           {align=center,text width=2.0cm,rounded corners=3pt, fill= magenta!10,draw= magenta!80,line width=0.3mm},   
    tuning_work/.style =           {align=center,text width=8.5cm,rounded corners=3pt, fill= magenta!10,draw= magenta!0,line width=0.3mm},          
}

\newcommand{\lstbg}[3][0pt]{{\fboxsep#1\colorbox{#2}{\strut #3}}}

\lstdefinelanguage{diff}{
  basicstyle=\ttfamily\small,
  morecomment=[f][\lstbg{red!20}]-,
  morecomment=[f][\lstbg{green!20}]+,
}

\lstdefinelanguage{diffpython}{
  language=diff,
  morekeywords={def, if, else, for, while, return, import, from, as, class, with, try, except, finally, raise, lambda, and, or, not, in, is, None, True, False},
  morecomment=[l]{\#},
  morestring=[b]",
  morestring=[b]',
}

\usepackage{xspace}
\newcommand{\ourmethod}{{\texttt{Mem-T}}\xspace} 
\newcommand{\ourtrain}{{\textsc{MoT-GRPO}}\xspace} 

\usepackage{xcolor}

\definecolor{ForestGreen}{RGB}{34,139,34}
\definecolor{myyellow}{RGB}{181, 181, 27}

\usepackage[table,xcdraw,usenames,dvipsnames]{xcolor}

\usepackage{amsmath}
\usepackage{amssymb}
\usepackage{mathtools}
\usepackage{amsthm}
\usepackage{fontawesome5} 

\usepackage[capitalize,noabbrev]{cleveref}

\usepackage{multirow}
\usepackage{makecell}
\usepackage{enumerate}
\usepackage{enumitem}
\usepackage{pifont}

\definecolor{mygrey}{gray}{0.4}

\usepackage[ruled,algo2e,vlined]{algorithm2e}
\SetKwInOut{Input}{Input}\SetKwInOut{Output}{Output}
\SetKwComment{Comment}{$\triangleright$\ }{}

\definecolor{darkgreen}{RGB}{30, 130, 30}
\definecolor{cream}{RGB}{253, 250, 242}
\renewcommand{\cmark}{\textcolor{darkgreen}{\ding{51}}} 
\renewcommand{\xmark}{\textcolor{red}{\ding{55}}}       

\newcommand{\stateNone}{\textcolor{lightgray}{\faTimes}} 
\newcommand{\stateFixed}{\textcolor{cyan}{\faTools}} 
\newcommand{\stateTrain}{\textcolor{red}{\faFire}}

\usepackage{listings}

\usepackage{ulem}
\usepackage{pifont}

\setheadertext{LUMIA Lab}

\correspondingemail{\emailicon\ ywyue25@stu.pku.edu.cn \quad $^\ddagger$ Corresponding Author.}

\githublink{https://github.com/yanweiyue/Mem-T} \huggingfacelink{https://huggingface.co/EdwinYue/Mem-T-4B}

\title{Mem-T: Densifying Rewards for Long-Horizon Memory Agents}
\setheadertitle{Mem-T: Densifying Rewards for Long-Horizon Memory Agents}

\author{%
  Yanwei Yue$^{1}$,
  Boci Peng$^{1}$, Xuanbo Fan$^{1}$, Jiaxin Guo$^{1}$, Qiankun Li$^{2\dagger}$, Yan Zhang$^{1\dagger}$\\
  $^1$Peking University
  $^2$Nanyang Technological University \\
  
}

\begin{document}

\begin{abstract}
Memory agents, which depart from predefined memory-processing pipelines by endogenously managing the processing, storage, and retrieval of memories, have garnered increasing attention for their autonomy and adaptability. However, existing training paradigms remain constrained: agents often traverse long-horizon sequences of memory operations before receiving sparse and delayed rewards, which hinders truly end-to-end optimization of memory management policies. To address this limitation, we introduce \ourmethod, an autonomous memory agent that interfaces with a lightweight hierarchical memory database to perform dynamic updates and multi-turn retrieval over streaming inputs. To effectively train long-horizon memory management capabilities, we further propose \ourtrain, a tree-guided reinforcement learning framework that transforms sparse terminal feedback into dense, step-wise supervision via memory operation tree backpropagation and hindsight credit assignment, thereby enabling the joint optimization of memory construction and retrieval.
Extensive experiments demonstrate that \ourmethod is \textbf{\ding{182} high-performing}, surpassing frameworks such as A-Mem and Mem0 by up to $14.92\%$, and \textbf{\ding{183} economical}, operating on a favorable accuracy-efficiency Pareto frontier and reducing inference tokens per query by $\sim24.45\%$ relative to GAM without sacrificing performance.
\end{abstract}

\maketitle


\section{Introduction}

\begin{wrapfigure}{r}{0.6\linewidth}
    \vspace{-0.8em}
    \centering
    \includegraphics[width=\linewidth]{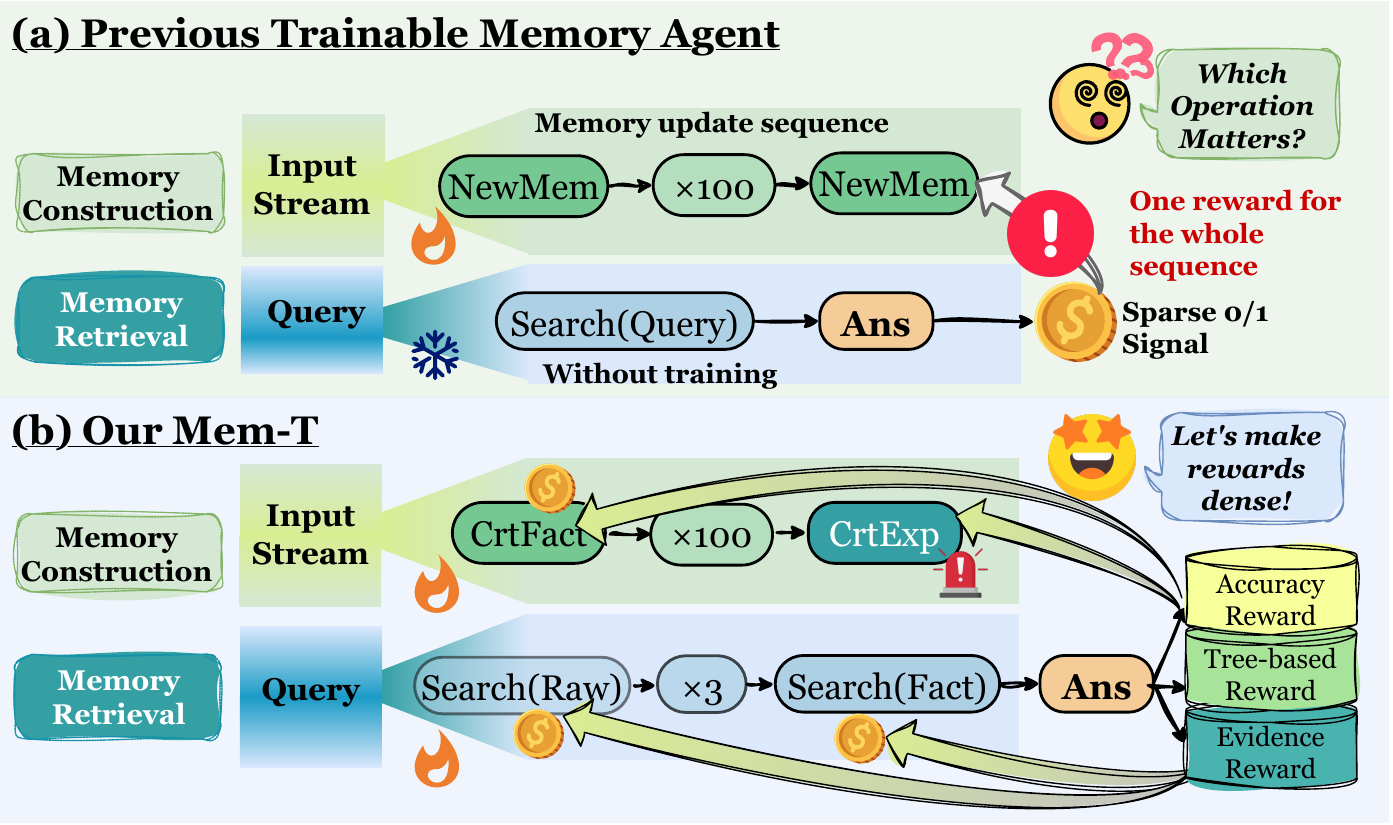}
    \caption{The paradigm comparison between the previous trainable memory agent and \ourmethod.}
    \label{fig:intro}
    \vspace{-1em}
\end{wrapfigure}

As Large Language Models (LLMs) rapidly evolve into powerful AI agents, they have achieved significant success across various fields~\citep{hong2023metagpt,wu2024autogen, qian2025toolrl, yang2025Qwen3technicalreport,xu2025comprehensivesurveydeepresearch}. 
However, constrained by the finite context windows of foundation models, AI agents face inherent challenges with long-term inconsistency~\citep{li2024long,liu2025comprehensivesurveylongcontext} and context forgetting during extended multi-turn interactions~\citep{ai2025memorybench,liu2025comprehensivesurveylongcontext}. 
As a promising frontier, memory systems dynamically construct and leverage memories from historical interactions~\citep{li2025memosoperatingmemoryaugmentedgeneration,fang2025lightmem}, thereby sustaining temporal coherence and long-term intelligence beyond finite context windows ~\citep{DBLP:journals/corr/abs-2509-12810,DBLP:conf/aaai/Zhao0XLLH24}, and have consequently emerged as a core component of modern agentic systems \citep{Chhikara2025mem0, zhang2025memevolve}.

Tracing the evolution of memory systems, early frameworks such as MemGPT~\citep{packer2023memgpt}, Mem0~\citep{Chhikara2025mem0}, and A-Mem~\citep{xuAMEMAgenticMemory2025} predominantly rely on hand-crafted prompts and heuristic rules to guide frozen LLMs in populating predefined memory structures. As a result, their performance is inherently bounded by the base model’s instruction-following capacity and rigid human priors, often leading to suboptimal outcomes~\citep{xiong2025memorymanagementimpactsllm,DBLP:conf/iclr/WuWYZCY25}. 
By contrast, recent approaches such as Memory-R1~\citep{yan2025memory}, Mem-$\alpha$~\citep{DBLP:journals/corr/abs-2509-25911}, and MemTool~\citep{lumer2025memtool} employ reinforcement learning (e.g., GRPO~\citep{shao2024deepseekmath}) to train LLMs into adaptive policies for dynamic memory curation and retrieval, commonly referred to as \emph{memory agents}. This shift constitutes a fundamental paradigm change, recasting memory management from static instruction adherence into a problem of adaptive policy optimization~\citep{hu2026memoryageaiagents}.

However, current paradigms for training memory agents remain fundamentally constrained by temporal credit assignment~\citep{pignatelli2024surveytemporalcreditassignment}, \textit{i.e.}, the challenge of attributing sparse and delayed rewards to causative actions along long-horizon memory operation sequences. This limitation is particularly acute in memory-centric tasks, where agents may execute hundreds of memory operations across $\sim$500 turns within million-token contexts before receiving a binary $0/1$ reward derived from sporadic QA accuracy signals~\citep{DBLP:conf/iclr/WuWYZCY25,tan2025membench}. Existing approaches fail to bridge this gap, as they indiscriminately propagate the sparse terminal reward across all memory operations without dense supervision or process-level attribution~\citep{yan2025memory,DBLP:journals/corr/abs-2509-25911}. Consequently, this extreme sparsity impedes effective optimization of the full memory operation trajectory. To put it more formally:


\begin{tcolorbox}[
    colback=cream,      
    colframe=black!80,  
    boxrule=0.8pt,      
    arc=4pt,            
    boxsep=2pt,         
    left=3pt, right=3pt, top=2pt, bottom=2pt, 
]
    \itshape 
   How can we implement a fully trainable memory agent framework that jointly optimizes memory construction and retrieval, supervised with dense rewards and accurate process-level attribution?
\end{tcolorbox}

To address this challenge, we introduce \ourmethod, a streamlined hierarchical memory agent optimized under a process-supervised, attribution-centric training paradigm termed \textbf{Memory Operation Tree-guided GRPO} (\ourtrain). Functionally, \ourmethod integrates three core capabilities: (i) formation and (ii) evolution operations that maintain and refine the hierarchical memory database over dynamic information streams, and (iii) a retrieval operation that conducts multi-turn, autonomous search to provide accurate memory clues.
To jointly optimize these components, \ourtrain employs a dual-track training mechanism integrating memory retrieval and construction.
To refine memory retrieval, it constructs multiple Memory operation Trees (MoT) to explore diverse trajectories, leveraging the branching topology to back-propagate sparse outcome rewards to intermediate nodes, thereby generating dense process-level signals and identifying critical search paths. 
To refine memory construction, the utility of the MoT is explicitly attributed back to source memory items via hindsight credit assignment, supervising the corresponding formation and evolution operations. 
This paradigm effectively mitigates reward sparsity and attribution ambiguity, rendering memory interactions both interpretable and learnable.
Our contributions can be summarized as:
\begin{itemize}[leftmargin=*,itemsep=-0em]
    \item \textbf{\textit{Unified Memory Framework.}} We propose \ourmethod, a streamlined memory management agent with a hierarchical architecture that integrates factual, experiential, and working memory, and agentically orchestrates the full lifecycle of memory operations.
    \item \textbf{\textit{Tree-Guided Optimization.}} We present \ourtrain, a memory operation tree-based paradigm that tackles temporal credit assignment via node-wise reward backpropagation and hindsight credit assignment. By transforming sparse terminal rewards into dense supervision for intermediate operations, it enables the joint optimization of memory formation, evolution, and retrieval.
    \item \textbf{\textit{Experimental Evaluation.}} Comprehensive evaluations on four memory benchmarks demonstrate that \ourmethod achieves state-of-the-art performance while maintaining a superior Pareto frontier, delivering up to $14.92\%$ F1 gains and reducing inference tokens per query by $\sim24.45\%$ compared with GAM and A-Mem baselines.

\end{itemize}

\section{Related Work}
\begin{wraptable}{r}{0.5\linewidth}
    \vspace{-0.8em}
    \centering
    \setlength{\tabcolsep}{3pt}
    \caption{Comparison of different memory agent systems. 
    \stateNone: Not included; 
    \stateFixed: Included but heuristic-based; 
    \stateTrain: Included and trainable.
    \textbf{Abbreviations}: Fact.=Factual Memory, Exp.=Experiential Memory, Work.=Working Memory, Form.=Memory Formation, Evol.=Memory Evolution, Retr.=Memory Retrieval, Proc. Attr.=Process Attribution.}
    \label{tab:comparison}
    \resizebox{\linewidth}{!}{%
    \begin{tabular}{l ccc ccc c}
    \toprule
    \multirow{2}{*}{\textbf{Method}} & \multicolumn{3}{c}{\textbf{Functions}} & \multicolumn{3}{c}{\textbf{Operations}} & \multirow{2}{*}{\shortstack{\textbf{Proc.}\\\textbf{Attr.}}} \\
    \cmidrule(lr){2-4} \cmidrule(lr){5-7}
     & Fact. & Exp. & Work. & Form. & Evol. & Retr. & \\
    \midrule
    MemAgent & \stateNone & \stateNone & \stateTrain & \stateTrain & \stateTrain & \stateNone & \stateNone \\
    Context-Folding & \stateNone & \stateNone & \stateTrain & \stateTrain & \stateFixed & \stateNone & \stateNone \\
    Memory-R1       & \stateTrain & \stateNone & \stateNone & \stateFixed & \stateTrain & \stateFixed & \stateNone \\
    Mem-$\alpha$  & \stateTrain & \stateNone & \stateTrain & \stateTrain & \stateTrain & \stateFixed & \stateNone \\
    MemSearcher & \stateTrain & \stateNone & \stateNone & \stateNone & \stateNone & \stateTrain & \stateNone \\
    LightSearcher & \stateNone & \stateTrain & \stateNone & \stateFixed & \stateNone & \stateTrain & \stateNone \\
    \midrule
    \ourmethod & \stateTrain & \stateTrain & \stateTrain & \stateTrain & \stateTrain & \stateTrain & \stateTrain \\
    \bottomrule
    \end{tabular}%
    }
    \vspace{-0.1em}
\end{wraptable}

\paragraph{Memory Agent Architectures.} 
In recent years, memory agents have advanced rapidly, evolving from heuristic-based systems such as MemoryBank~\citep{zhong2023memorybankenhancinglargelanguage} and MemGPT~\citep{packer2023memgpt} to more agentic architectures, including Mem0~\citep{Chhikara2025mem0}, MemOS~\citep{li2025memosoperatingmemoryaugmentedgeneration}, and A-Mem~\citep{xuAMEMAgenticMemory2025}. 
\textbf{Functionally}, prior work spans three categories: \textit{(I) Factual Memory}, preserving declarative knowledge for long-term consistency~\citep{zhong2023memorybankenhancinglargelanguage}; \textit{(II) Experiential Memory}, distilling experience from trajectories to support continual self-improvement~\citep{DBLP:conf/aaai/Zhao0XLLH24}; and \textit{(III) Working Memory}, managing dynamic context for ongoing tasks~\citep{wuReSumUnlockingLongHorizon2025}. \textbf{Operationally}, the memory lifecycle comprises \textit{(I) Formation}, transforming raw context into high-value memory; \textit{(II) Evolution}, integrating new insights with existing memory store; and \textit{(III) Retrieval}, performing accurate retrieval from the memory base. As shown in \Cref{tab:comparison}, our \ourmethod, despite its streamlined design, spans all three functional classes and operational stages.

\vspace{-0.1em}
\paragraph{Reinforcement Learning for Memory Agents.}
As memory systems scale in complexity, the efficacy of foundation models in managing memory increasingly becomes the primary performance bottleneck. Consequently, reinforcement learning (RL) has emerged as a central paradigm for endowing LLMs with adaptive memory management capabilities~\citep{hu2026memoryageaiagents}. Current research spans a broad spectrum, from short-term working memory to long-term factual and experiential memory.
\textit{Working Memory.} RL has been used to enable agents to autonomously manage execution context within a single task~\citep{yu2025memagent, chen2025iterresearchrethinkinglonghorizonagents}, particularly in settings such as deep research and web browsing~\citep{zhou2025mem1learningsynergizememory, DBLP:journals/corr/abs-2510-11967, ye2025agentfold}.
\textit{Long-term Factual Memory.} Prior work targets different stages of memory management: Memory-R1~\citep{yan2025memory} emphasizes memory evolution, Mem-$\alpha$~\citep{DBLP:journals/corr/abs-2509-25911} addresses both formation and evolution, and MemSearcher~\citep{yuanMemSearcherTrainingLLMs2025} focuses on training agents to exploit retrieval tools.
\textit{Long-term Experiential Memory.} Methods such as LightSearcher~\citep{lan2025lightsearcherefficientdeepsearchexperiential} and MemRL~\citep{zhang2026memrlselfevolvingagentsruntime} improve the acquisition, refinement, and reuse of skills over time.
Despite these advances, RL-based approaches remain limited by sparse rewards and temporal credit assignment in long-horizon settings, hindering effective optimization across the full memory construction and utilization pipeline, as shown in \Cref{tab:comparison}.

\begin{figure*}[!t]
    \centering
    \includegraphics[width=1\linewidth]{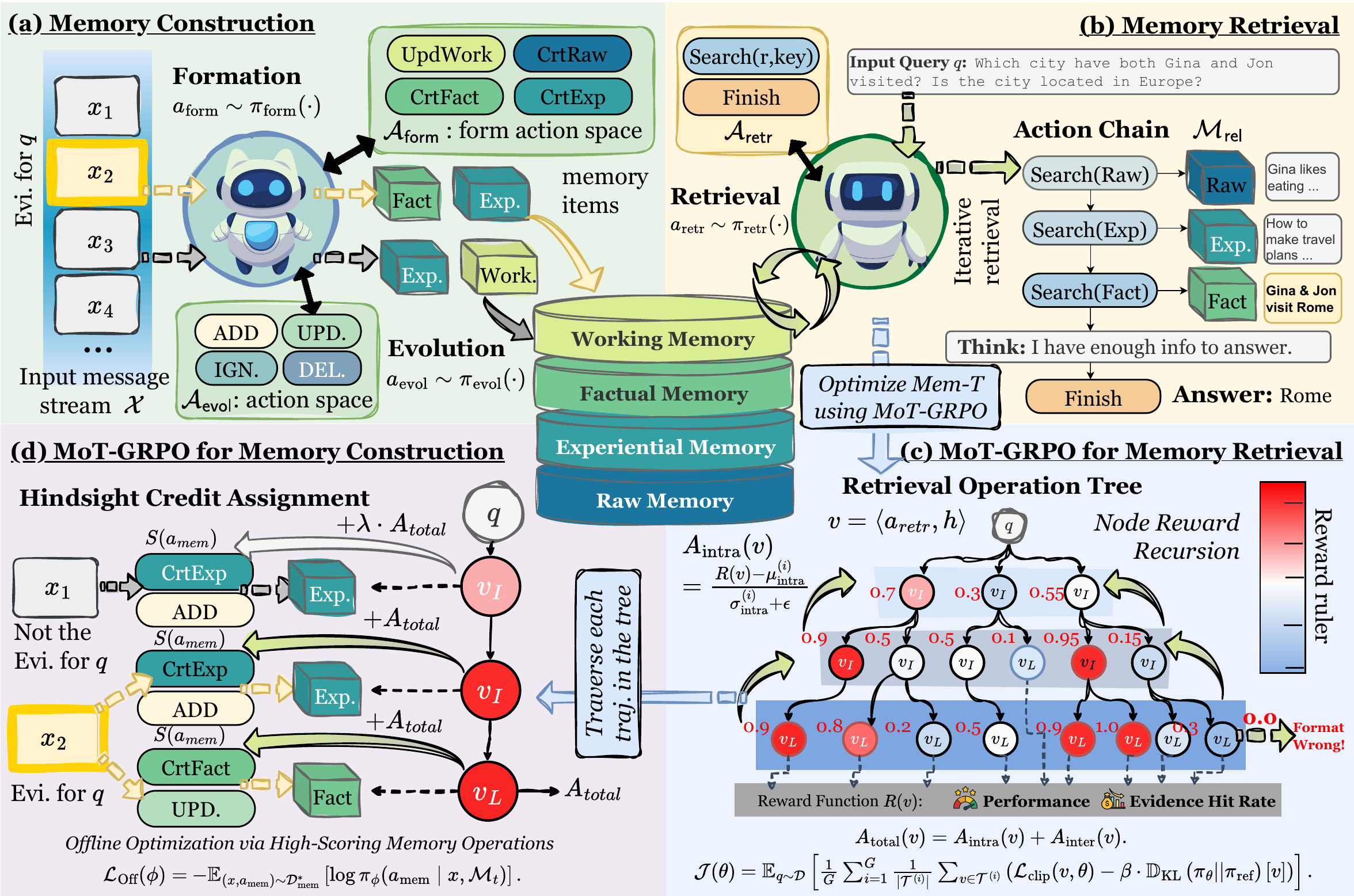}
    \caption{The overall framework of our proposed \ourmethod.}
    \label{fig:main}
    \vspace{-0.1em}
\end{figure*}

\vspace{-0.1em}
\section{Method}
\vspace{-0.1em}
\subsection{Mem-T Workflow}
\paragraph{Hierarchical Memory Definition.}
We consider the agent interacting with a continuous information stream $\mathcal{X} = \{x_1, x_2, \dots, x_T\}$. 
At each time step $t$, corresponding to the processing of the current chunk $x_t$, the system maintains a hierarchical memory state $\mathcal{M}_t$:
\begin{equation}
\mathcal{M}_t = \langle \mathcal{M}^{\text{work}}_t, \mathcal{M}^{\text{fact}}_t, \mathcal{M}^{\text{exp}}_t,  \mathcal{M}^{\text{raw}}_t\rangle.
\end{equation}
Within this hierarchy, \textbf{Working Memory} ($\mathcal{M}^{\text{work}}_t$) iteratively updates a concise summary at each step, maintaining within-episode coherence.
The long-term memory consists of three modules: \textbf{Factual Memory} ($\mathcal{M}^{\text{fact}}_t$) stores declarative knowledge, \textbf{Experiential Memory} ($\mathcal{M}^{\text{exp}}_t$) captures procedural knowledge, and \textbf{Raw Memory} ($\mathcal{M}^{\text{raw}}_t$) archives raw data across sessions. Formally, we have:

\begin{equation}
\begin{aligned}
    \mathcal{M}^{\text{fact}}_t &= \{ m^{\text{fact}}_i \mid m^{\text{fact}}_i = (f_i, t^{\text{start}}_i, t^{\text{end}}_i) \}_{i=1}^{N_f}, \\
    \mathcal{M}^{\text{exp}}_t &= \{ m^{\text{exp}}_j \mid m^{\text{exp}}_j = (e_j, t^{\text{start}}_j, t^{\text{end}}_j) \}_{j=1}^{N_e}, \\
    \mathcal{M}^{\text{raw}}_t &= \{ m^{\text{raw}}_l \mid m^{\text{raw}}_l = (x_l, t^{\text{raw}}_l) \}_{l=1}^{t}, \\
\end{aligned}
\end{equation}
where each $m^{(\cdot)}$ represents an atomic memory unit. Specifically, $f_i$ and $e_j$ represent concrete facts and strategies, respectively, bound by validity time windows $[t^{\text{start}}, t^{\text{end}}]$. 
\vspace{-0.5em}

\paragraph{Memory Operation Pipeline.}
Building upon this hierarchical memory, we formulate the agent’s interaction as a dual-track decision process, comprising continuous memory construction and on-demand memory utilization.

\noindent \textbf{Phase I: Continuous Memory Construction.}
As the agent processes the input stream $x_t$, it proactively constructs new memory candidates via the memory formation policy $\pi_{\text{form}}$. 
This policy scans the raw input to identify salient information and operates on the formation action space $\mathcal{A}_{\text{form}}=\{\texttt{CrtFact}, \texttt{CrtExp}, \texttt{CrtRaw}, \texttt{UpdWork}\}$. Here, \texttt{CrtFact}, \texttt{CrtExp}, and \texttt{CrtRaw} extract atomic declarative facts, procedural strategies, and raw data, respectively, while \texttt{UpdWork} updates the session-level working summary.
Formally, the formation process is defined as:
\begin{equation}
\begin{aligned}
a_{\text{form}} \sim \pi_{\text{form}}(\cdot | x_t, \mathcal{M}^{\text{work}}_t), \quad a_{\text{form}} \subseteq \mathcal{A}_{\text{form}}, \\
\mathcal{M}_t^{\text{cand}} = \{ m \mid m \leftarrow \text{Execute}(op), \forall op \in a_{\text{form}} \},
\end{aligned}
\end{equation}
where $\mathcal{M}_t^{\text{cand}}$ denotes the set of candidate memories extracted from $x_t$. For each candidate $m \in \mathcal{M}_t^{\text{cand}}$, the memory evolution policy $\pi_{\text{evol}}$ integrates it into $\mathcal{M}_{t}$.
Specifically, the policy considers memories in $\mathcal{M}_{t}$ that are relevant to $m$, and samples an evolution action $a_{\text{evol}} \sim \pi_{\text{evol}}(\cdot \mid m, \mathcal{M}_t)$ from the action space $\mathcal{A}_{\text{evol}}=\{\texttt{ADD}, \texttt{UPDATE}, \texttt{DELETE}, \texttt{IGNORE}\}$.
Collectively, these actions define the set of memories to be added ($\Delta^+$) and removed ($\Delta^-$) from the memory store:
\begin{equation}
\small
\begin{aligned}
    \Delta^+ &= \{ m | a_{\text{evol}} = \texttt{ADD} \} \cup \{ m_{\text{refined}} | a_{\text{evol}} = \texttt{UPDATE} \}, \\
    \Delta^- &= \{ m_{\text{target}} | a_{\text{evol}} = \texttt{DELETE} \} \cup \{ m_{\text{old}} | a_{\text{evol}} = \texttt{UPDATE} \}.
\end{aligned}
\end{equation}
Consequently, the memory store is updated accordingly:
\begin{equation}
\mathcal{M}_{t+1} = (\mathcal{M}_t \setminus \Delta^-) \cup \Delta^+.
\end{equation}

\noindent \textbf{Phase II: On-Demand Memory Retrieval.}
Based on the constructed memory store $\mathcal{M}_{t}$, when a query $q$ arises, the agent employs a multi-turn retrieval to respond. 
During this process, the memory retrieval policy $\pi_{\text{retr}}$ selects actions from the retrieval action space $\mathcal{A}_{\text{retr}}$, which includes queries for each memory module and a terminal signal:
\begin{equation}
\mathcal{A}_{\text{retr}} = \{ \texttt{Search}(r,\text{key},\text{topk}) \mid r \in \mathcal{M}_{t} \} \cup \{ \texttt{Finish} \},
\end{equation}
where $r$ is the memory type to be retrieved, $\text{key}$ is the retrieval query. 
Unlike single-step retrieval, $\pi_{\text{retr}}$ operates as a sequential decision policy. 
At each step $k$, conditioned on the query $q$ and the history context $h_{k-1}$, which consists of the retrieved relevant memory set $\mathcal{M}^{\text{rel}}_{k-1}$ and reasoning state $z_{k-1}$, the policy samples an action $a_k$:
\begin{equation}
\label{eq:retract}
a_k \sim \pi_{\text{retr}}(\cdot \mid q, \mathcal{M}_t, h_{k-1}), \quad a_k \in \mathcal{A}_{\text{retr}}.
\end{equation}
This iterative process accumulates the relevant memory set $\mathcal{M}^{\text{rel}}$ by aggregating the observations from each search step. 
Finally, the loop terminates when the policy selects the \texttt{Finish} action, signaling that the gathered information is sufficient to support the final answer $y \sim P_{\theta}(\cdot \mid q, \mathcal{M}^{\text{rel}})$.

\subsection{MoT-GRPO for Memory Retrieval}
In long-horizon scenarios, memory operation chains become extremely long, making credit assignment and reward sparsity major challenges. To address these issues, we propose Memory Operation Tree GRPO (\ourtrain), inspired by prior RL methods~\citep{shao2024deepseekmath,ji2025treesearchllmagent}.
\label{sec:MoTGRPO}
\paragraph{Memory Operation Tree Construction.}
In the retrieval phase, to achieve efficient rollout generation while obtaining dense intermediate signals, we employ an \textbf{Iterative Branching Rollout} to construct the Memory Operation Tree(MoT). 
Formally, we define a node in MoT as a tuple $v = \langle a_{retr}, h \rangle$, representing a specific operation $a_{retr} \in \mathcal{A}_{\text{retr}}$ and the reasoning context $h$.

For each query, we initialize an ensemble of $G$ independent MoTs $\{ \mathcal{T}^{(i)}_0 \}_{i=1}^G$.
Each tree $\mathcal{T}^{(i)}_0$ initially contains a single seed trajectory $\tau^{(i)}$, obtained by a full rollout from the root state $(q, \mathcal{M}_t, h_{0}=\emptyset)$:
\begin{equation}
\small
\begin{aligned}
\mathcal{T}^{(i)}_0 &= \{ \tau^{(i)} \}, \quad i=1, \dots, G \\
\tau^{(i)} &= (v^{(i)}_1, v^{(i)}_2, \dots, v^{(i)}_{L_i}), \text{where } v^{(i)}_k = \langle a^{(i)}_k, h^{(i)}_k \rangle.
\end{aligned}
\end{equation}

Subsequently, we iteratively densify each $\mathcal{T}^{(i)}$ over $M$ expansion rounds.
In each expansion round $j \in \{1, \dots, M\}$, we stochastically sample $N_v$ non-terminal pivot nodes $\{v^*_n\}_{n=1}^{N_v}$ from each tree $\mathcal{T}^{(i)}_{j-1}$.
For each node $v^*$ and its corresponding context history $h_{v^*}$, the policy executes a new rollout to generate a branch trajectory $\tau_{\text{branch}}$:
\begin{equation}
\begin{aligned}
\tau_{\text{branch}} &\sim \pi_{\text{retr}}(\cdot \mid q, \mathcal{M}_t, h_{v^*}), \\
\tau_{\text{new}} &= \text{Path}(v^*) \oplus \tau_{\text{branch}}.
\end{aligned}
\end{equation}
The newly generated trajectories are then grafted onto the tree, updating its state to $\mathcal{T}^{(i)}_j$. After $M$ rounds, this process yields a final ensemble of $G$ MoTs $\{ \mathcal{T}^{(i)}_M \}_{i=1}^G$.

\paragraph{Node-wise Reward Backpropagation.}
Instead of relying solely on sparse terminal rewards, we assign a dense reward $R(v)$ to every node $v$, synthesizing immediate retrieval quality with expected future success. Formally, for a node $v$ with retrieved memories $\mathcal{M}^{\text{rel}}_v$, we define the reward as:
\begin{equation}
R(v) = \mathbb{I}_{\text{fmt}}(v) \cdot \left( \alpha \cdot \text{Evid}(v) +  \text{Perform}(v) \right)
\end{equation}
Here, $\mathbb{I}_{\text{fmt}}(v)$ serves as a binary validity mask ensuring syntactic correctness of tool invocations; $\text{Evid}(v)$ measures the immediate evidence density, calculated as the proportion of ground-truth evidence retrieved in $\mathcal{M}^{\text{rel}}_v$; and $\text{Perform}(v)$ denotes the expected terminal performance of node $v$. For a leaf node, it is defined as the answer quality measured by the F1 score or accuracy. For an internal node, it is computed as the average $\text{Perform}(\cdot)$ over all its child nodes $\mathrm{Ch}(v)$:
$$
\text{Perform}(v) = 
\begin{cases}
\mathrm{F1}(v), & v \in \mathcal{V}_{\text{leaf}}, \\
\frac{1}{|\mathrm{Ch}(v)|} \displaystyle\sum_{u \in \mathrm{Ch}(v)} \mathrm{Perform}(u), 
& \text{otherwise}.
\end{cases}
$$

This formulation ensures that high-reward nodes should adhere to valid formats, retrieve relevant evidence, and lead to high-quality outcomes.

\paragraph{Dual-Scale Advantage Estimation.}
To enable tree-based credit assignment, we perform grouped advantage estimation at both the intra-tree and inter-tree levels. The \textit{Intra-Tree Advantage} $A_{\text{intra}}(v)$ evaluates the relative quality of nodes within the same tree. For a node $v$ in tree $\mathcal{T}^{(i)}$, we standardize $R(v)$ using the mean $\mu_{\text{intra}}^{(i)}$ and standard deviation $\sigma_{\text{intra}}^{(i)}$ derived from that specific tree:
\begin{equation}
A_{\text{intra}}(v) = \frac{R(v) - \mu_{\text{intra}}^{(i)}}{\sigma_{\text{intra}}^{(i)} + \epsilon}
\end{equation}
Simultaneously, to capture each node’s global advantage, we compute the \textit{Inter-Tree Advantage} $A_{\text{inter}}(v)$ against the global mean $\mu_{\text{global}}$ and standard deviation $\sigma_{\text{global}}$ across the entire ensemble $\{ \mathcal{T}^{(i)} \}_{i=1}^G$:
\begin{equation}
A_{\text{inter}}(v) = \frac{R(v) - \mu_{\text{global}}}{\sigma_{\text{global}} + \epsilon}
\end{equation}
The final advantage $A_{\text{total}}(v)$ balances these perspectives:
\begin{equation}
A_{\text{total}}(v) = A_{\text{intra}}(v) + A_{\text{inter}}(v).
\end{equation}
Through this dual-scale design, the intra-tree advantage supports reliable local comparisons sharing similar contexts and effective credit assignment to identify nodes that critically influence the final outcome. Meanwhile, inter-tree advantages encourage cross-tree competition, guiding the optimization toward globally high-quality solutions.

\paragraph{Optimization Objective.}
Following the GRPO paradigm, we directly utilize the dual-scale advantage $A_{\text{total}}(v)$ to optimize the retrieval policy $\pi_{\theta}$ by maximizing:
\begin{equation}
\small
\begin{aligned}
\mathcal{J}(\theta) = \mathbb{E}_{q \sim \mathcal{D}} \left[ \frac{1}{G} \sum_{i=1}^G \frac{1}{|\mathcal{T}^{(i)}|} \sum_{v \in \mathcal{T}^{(i)}} \left( \mathcal{L}_{\text{clip}}
- \beta \mathbb{D}_{\text{KL}} \left( \pi_{\theta} || \pi_{\text{ref}} \right) \right) \right]
\end{aligned}
\end{equation}
where $\pi_{\text{ref}}$ constrains the update via the KL penalty coefficient $\beta$. The core term $\mathcal{L}_{\text{clip}}$ applies standard PPO clipping to the probability ratio $\rho_{v,t}(\theta) = \pi_{\theta}(a_{v,t} | \cdot) / \pi_{\theta_{\text{old}}}(a_{v,t} | \cdot)$:
\begin{equation}
\small
\mathcal{L}_{\text{clip}} = \min \left(\rho_{v,t}(\theta) A_{\text{total}}(v), \text{clip}(\rho_{v,t}(\theta), 1\pm\epsilon) A_{\text{total}}(v) \right)
\end{equation}

\subsection{MoT-GRPO for Memory Construction}
Unlike retrieval, memory construction spans hundreds of steps with rewards delayed until downstream queries, and its quality is irrelevant to most queries, resulting in severe credit assignment ambiguity.
To address this, we propose \textbf{Hindsight Credit Assignment}, which back-propagates advantage signals from downstream retrieval trajectories to upstream construction actions.

\paragraph{Hindsight Credit Assignment.}

Let $a_{\text{mem}}$ be a memory operation processing source turns $\mathcal{X}_{\text{src}}$ to produce a memory entry $m$. For a query $q$ with ground-truth evidence $\mathcal{X}_{\text{evi}}^{q}$, we define the hindsight score $S(a_{\text{mem}})$ by aggregating advantages $A_{\text{total}}(v_L)$ from terminal leaf nodes $v_L \in \mathcal{V}_{\text{leaves}}$:
\begin{equation}
S(a_{\text{mem}}) = \frac{1}{|\mathcal{V}_{\text{leaves}}|} \sum_{v_L \in \mathcal{V}_{\text{leaves}}} A_{\text{total}}(v_L) \cdot \varrho(a_{\text{mem}}, v_L)
\end{equation}
The credit coefficient $\varrho$ integrates two distinct signals:
\begin{equation}
\varrho(a_{\text{mem}}, v_L) = \underbrace{\mathbb{I}(\mathcal{X}_{\text{src}} \cap \mathcal{X}_{\text{evi}}^{q} \neq \emptyset)}_{\text{Evidence Alignment Gate}} + \lambda \cdot \underbrace{\mathbb{I}(m \in \mathcal{M}_{v_L})}_{\text{Retrieval Trace Gate}}
\end{equation}
The \textit{Evidence Alignment Gate} attributes credit by linking the construction quality of ideal evidence turn $\mathcal{X}_{\text{evi}}^{q}$ to answer accuracy. It posits that successful reasoning is fundamentally rooted in the effective transformation of ground-truth evidence into memory. Thus, the advantage of a final answer serves as a proxy to evaluate the construction of these pivotal source turns. Conversely, the \textit{Retrieval Trace Gate} (weighted by $\lambda=0.1$) captures the empirical utility of $m$ retrieved within the actual retrieval tree. It recognizes that any memory entry $m$ involved in the terminal path $\mathcal{M}_{v_L}$ objectively modulates the model's decision-making, rewarding the construction process for its functional contribution to the successful trajectory. Notably, in the absence of ground-truth evidence, the mechanism naturally relies on the \textit{Retrieval Trace Gate}, maintaining robust generalization across diverse datasets.

\begin{table*}[!ht]
    \centering
    \caption{Performance comparison on the LoCoMo benchmark, with F1 and BLEU-1 as the evaluation metrics. $^{\dagger}$: The GAM paper recommends gpt-4o-mini; we also reproduced it using Qwen3-4B for a fair comparison. $^{\ddagger}$: As Memory-R1 is not open-source, we faithfully report the results provided in their original paper.}
    \label{tab:LoCoMo_results}
    \resizebox{\textwidth}{!}{%
    \begin{tabular}{llcccccccccc}
        \toprule
        \multirow{2}{*}{\textbf{Method}} & \multirow{2}{*}{\textbf{Base LLM}} & \multicolumn{2}{c}{\textbf{Single-Hop}} & \multicolumn{2}{c}{\textbf{Multi-Hop}} & \multicolumn{2}{c}{\textbf{Temporal}} & \multicolumn{2}{c}{\textbf{Open Domain}} & \multicolumn{2}{c}{\textbf{Overall}} \\
        \cmidrule(lr){3-4} \cmidrule(lr){5-6} \cmidrule(lr){7-8} \cmidrule(lr){9-10} \cmidrule(lr){11-12}
         & & F1$\uparrow$ & B1$\uparrow$ & F1$\uparrow$ & B1$\uparrow$ & F1$\uparrow$ & B1$\uparrow$ & F1$\uparrow$ & B1$\uparrow$ & F1$\uparrow$ & B1$\uparrow$ \\
        \midrule
        \multicolumn{12}{l}{\textit{Training-free Methods}} \\
        VANILLA & Qwen3-4B & 40.68 & 31.54 & 23.23 & 16.76 & 18.97 & 13.42 & 13.87 & 10.70 & 31.50 & 23.94 \\
        RAG  & Qwen3-4B & 49.45 & 44.94 & 23.50 & 17.13 & 43.07 & 37.35 & 20.23 & 14.94 & 41.59 & 36.45 \\
        MemGPT & Qwen3-4B & 14.00 & 11.77 & 16.68 & 13.99 & 12.56 & 10.94 & 11.61 & 9.16 & 14.05 & 11.84 \\
        MemoryBank & Qwen3-4B & 26.65 & 17.72 & 25.52 & 19.44 & 9.15 & 7.44 & 16.42 & 12.39 & 22.34 & 15.66 \\
        Mem0   & Qwen3-4B & 47.28 & 40.72 & 35.40 & 27.36 & 46.84 & 39.48 & 26.64 & 21.04 & 43.71 & 36.78 \\
        MemoryOS  & Qwen3-4B & 48.35 & 42.57 & 35.24 & 27.30 & 40.98 & 32.68 & 22.08 & 17.93 & 42.83 & 36.26 \\
        LightMem  & Qwen3-4B & 43.78 & 38.84 & 30.78 & 25.80 & 44.71 & 40.72 & 18.93 & 14.42 & 40.01 & 35.27 \\
        A-Mem & Qwen3-4B & 44.62 & 38.26 & 27.24 & 21.07 & 43.85 & 35.97 & 15.40 & 12.71 & 39.43 & 33.04 \\
        GAM  & Qwen3-4B  & 32.23 & 25.54 & 32.23 & 28.66 & 26.26 & 22.52 & 18.45 & 14.47 & 30.17 & 24.81 \\
        \color{gray} GAM$^{\dagger}$  & \color{gray} gpt-4o-mini$^{\dagger}$ & \color{gray} 57.75 & \color{gray}52.10 & \color{gray}42.29 & \color{gray}34.44 & \color{gray}59.45 & \color{gray}53.11 & \color{gray}29.73 & \color{gray}24.74 & \color{gray}53.48 & \color{gray}47.33 \\
        \midrule
        \multicolumn{12}{l}{\textit{Trained Methods}} \\
        MEM1  & MEM1-7B & 27.48 & 22.10 & 18.98 & 15.56 & 30.52 & 23.48 & 14.21 & 11.43 & 25.68 & 20.50 \\
        MemAgent & MemAgent-14B & 35.86 & 29.64 & 27.86 & 22.72 & 37.93 & 31.85 & 20.31 & 16.47 & 33.82 & 27.97 \\
        Memory-R1-PPO$^{\ddagger}$  & Mem-R1-8B$^{\ddagger}$ & 32.52 & 24.47 & 26.86 & 23.47 & 41.57 & 26.11 & \underline{45.30} & \underline{39.18} & 34.08 & 25.54 \\
        Memory-R1-GRPO$^{\ddagger}$  & Mem-R1-8B$^{\ddagger}$ & 35.73 & 27.70 & 35.65 & 30.77 & 49.86 & 38.27 & \textbf{47.42} & \textbf{41.24} & 39.25 & 31.21 \\
        \midrule
        \multicolumn{12}{l}{\textit{Our Method \ourmethod}} \\
        \textit{w/o} training & Qwen3-4B & 53.97 & 49.15 & 38.44 & 31.70 & 53.99 & 48.08 & 26.44 & 23.37 & 49.38 & 44.11 \\
        with GRPO & Qwen3-4B & \underline{59.43} & \underline{54.65} & 38.40 & 30.51 & \underline{60.78} & \underline{56.10} & 23.46 & 20.16 & \underline{53.56} & \underline{48.33} \\
        with \ourtrain & Qwen3-4B & \textbf{63.75} & \textbf{57.95} & \textbf{45.09} & \textbf{36.58} & \textbf{65.13} & \textbf{60.12} & 32.97 & 28.94 & \textbf{58.65} & \textbf{52.63} \\
        \bottomrule
    \end{tabular}%
    }
    \vspace{-0.5em}
\end{table*}

\paragraph{Policy Refinement.}
To optimize memory construction policies, we employ rank-based sampling to curate a high-quality training dataset $\mathcal{D}_{\text{mem}}^*$. We first discard trajectories with invalid tool invocations. Subsequently, we rank all candidate actions by their hindsight score $S(a_{\text{mem}})$ and retain only the top $50\%$ percentile within each operation category. 
Finally, treating $\mathcal{D}_{\text{mem}}^*$ as a collection of expert demonstrations, we train the policies $\pi_{\theta}$ (encompassing $\pi_{\text{form}}$ and $\pi_{\text{evol}}$) to maximize the log-likelihood of these selected actions:
\begin{equation}
\mathcal{L}_{\text{Off}}(\theta) = - \mathbb{E}_{(x, a_{\text{mem}}) \sim \mathcal{D}_{\text{mem}}^*} \left[ \log \pi_{\theta}(a_{\text{mem}} \mid x, \mathcal{M}_t) \right].
\end{equation}
This offline optimization effectively distills the ``hindsight wisdom" derived from the downstream MoT-GRPO search trees into the forward-looking memory construction policy.

\vspace{-0.3em}
\section{Experiments}
\vspace{-0.3em}

\subsection{Experimental Setup}
\vspace{-0.3em}
\paragraph{Evaluation and Benchmarks.} We evaluate the proposed framework across four challenging long-context benchmarks, including LoCoMo~\citep{maharana2024evaluating}, LongMemEval~\citep{DBLP:conf/iclr/WuWYZCY25}, HotpotQA~\citep{yang2018hotpotqadatasetdiverseexplainable}, and NarrativeQA~\citep{kočiský2017narrativeqareadingcomprehensionchallenge}. LoCoMo and LongMemEval focus on long-term conversational question answering.
Following Memory-R1~\citep{yan2025memory}, we use the same training data configuration by splitting the LoCoMo dataset into a 1:1:8 train/validation/test split to ensure a fair comparison. The remaining three benchmarks are treated as out-of-domain datasets to evaluate the generalization ability of our method.
Specifically, for HotpotQA, following~\citep{yu2025memagent,yanGeneralAgenticMemory2025}, we construct long-context inputs by concatenating the gold supporting documents with 400 irrelevant Wikipedia documents. More details about the dataset are in \Cref{app:dataset}.
\vspace{-1em}

\paragraph{Baselines.} We compare \ourmethod against thirteen baselines, categorized into two groups: \textbf{(I) Training-free Methods:} This group includes memory-free approaches, such as vanilla long-LLM and retrieval-augmented generation (RAG)~\citep{DBLP:conf/nips/RAG2020}, as well as memory-based methods, including MemGPT~\citep{packer2023memgpt}, MemoryBank~\citep{zhong2023memorybankenhancinglargelanguage}, Mem0~\citep{Chhikara2025mem0}, LightMem~\citep{fang2025lightmem}, A-Mem~\citep{xuAMEMAgenticMemory2025}, and GAM~\citep{yanGeneralAgenticMemory2025}. \textbf{(II) Training-based Methods:} This group includes MemAgent~\citep{yu2025memagent} and Mem1~\citep{zhou2025mem1learningsynergizememory}, which primarily focus on working memory, and Memory-R1~\citep{yan2025memory} and Mem-$\alpha$~\citep{DBLP:journals/corr/abs-2509-25911}, which are designed to mainly enhance factual memory. For all the baselines, official implementations and released parameters are used when available.

\paragraph{Implementation Details.} We select LLM backbones of varying sizes, including Qwen3-4B and Qwen3-8B~\citep{yang2025Qwen3technicalreport}. All methods use BGE-M3 as the embedding model~\citep{chen2025m3embedding}.
During training with \ourtrain, we generate three trees for each query ($G=3$), with a maximum tree depth of $4$. In each expansion round, we select three nodes ($N_v=3$) for branch expansion. The training for memory retrieval is conducted for $200$ steps. And the training for memory construction is based on a dataset containing $10k$ memory operations. 
At inference time, \ourmethod is allowed up to $6$ reasoning steps. All retrieval operations default to returning the top-5 most similar items. 
More training setup and parameter configurations are listed in \Cref{app:imple}.

\subsection{Main Results}
\paragraph{High Performance.}
As shown in~\Cref{tab:LoCoMo_results} and~\Cref{tab:LoCoMo_results_8B}, \ourmethod achieves substantially better performance on the LoCoMo benchmark than both training-free and training-based baselines. 
When using Qwen3-4B and Qwen3-8B, \ourmethod improves F1 by $14.92$ ($34.13\%\uparrow$) and $14.55$ ($33.08\%\uparrow$), respectively. 
Even without training, the hierarchical and highly agentic memory system of \ourmethod achieves superior performance, improving F1 by $5.67$ ($12.97\%\uparrow$) compared to other methods.
Moreover, \ourtrain further strengthens the LLM's memory management capability compared to the training-free and the GRPO baseline, yielding additional F1 gains of $9.27$ ($18.77\%\uparrow$) and $5.09$ ($9.50\%\uparrow$). 
These results demonstrate that the joint retrieval and construction training with dense rewards in \ourtrain is better suited for long-horizon memory agents. 
Notably, GAM, the SOTA memory system, exhibits an F1 gap of $23.31$ when switching its backbone from gpt-4o-mini to Qwen3-4B, highlighting the importance of systematically improving model-level memory management capabilities.

\begin{wraptable}{r}{0.6\linewidth}
    \vspace{-0.8em}
    \centering
    \caption{Evaluation results on OOD benchmarks (HotpotQA, LongMemEval, NarrativeQA). All methods, except MEM1, which uses the 7B model trained in the original paper, are implemented with models based on Qwen3-4B.}
    \label{tab:ood_results}
    \resizebox{\linewidth}{!}{%
    \begin{tabular}{lcccc}
        \toprule
        \multirow{2}{*}{\textbf{Method}} & \textbf{HotpotQA} & \textbf{LongMemEval} & \textbf{NarrativeQA} & \multirow{2}{*}{\textbf{Avg.}} \\
         & F1$\uparrow$ & Acc$\uparrow$ & F1$\uparrow$ & \\
        \midrule
        \multicolumn{5}{l}{\textit{Training-free Methods}} \\
        VANILLA     & 21.89 & 38.80 & 18.09 & 26.26 \\
        RAG         & 50.13 & 56.60 & 21.17 & 42.63 \\
        MemGPT      & 18.24 & 23.00 & 8.39  & 16.54 \\
        MemoryBank  & 16.90 & 26.20 & 9.65  & 17.58 \\
        A-Mem       & 30.46 & 61.30 & 25.18 & 38.98 \\
        Mem0        & 31.96 & 53.60 & 27.63 & 37.73 \\
        MemoryOS    & 26.86 & 46.80 & 23.45 & 32.37 \\
        LightMem    & 38.62 & \underline{63.10} & 16.78 & 39.50 \\
        GAM         & 52.98 & 61.80 & 28.32 & \underline{47.70} \\
        \midrule
        \multicolumn{5}{l}{\textit{Trained Methods}} \\
        MEM1        & 55.36 & 19.00 & 13.49 & 29.28 \\
        Mem-$\alpha$ & \underline{58.80} & 52.00 & \underline{28.56} & 46.45 \\
        \ourmethod    & \textbf{66.35} & \textbf{65.80} & \textbf{30.29} & \textbf{54.15} \\
        \bottomrule
    \end{tabular}%
    }
    \vspace{-0.1em}
\end{wraptable}

\paragraph{Cross-domain generalization.}
To evaluate whether the memory management capabilities learned by \ourtrain can transfer across tasks, we assess the performance of \ourmethod on three out-of-domain tasks. As shown in ~\Cref{tab:ood_results}, baselines such as LightMem achieve suboptimal performance on LongMemEval but fail to generalize to other benchmarks, trailing \ourmethod by $27.73$ and $13.51$ on HotpotQA and NarrativeQA, respectively. Training-based MEM-1 performs well on the in-domain QA benchmark HotpotQA, outperforming training-free methods by $2.38$, but suffers substantial degradation on benchmarks that emphasize long-horizon dialogue understanding, underperforming \ourmethod by $46.8$ and $16.8$. In contrast, \ourmethod learns effective memory management strategies through training on LoCoMo and achieves SOTA performance across all three out-of-domain benchmarks, with an average improvement of $6.45(13.52\%\uparrow)$ over other methods. Notably, \ourmethod generalizes well from long-horizon dialogue to the QA setting of HotpotQA, outperforming other approaches by $7.55$.

\begin{figure*}[!t]
    \centering
    \begin{minipage}{0.49\linewidth}
        \centering
        \includegraphics[width=\linewidth]{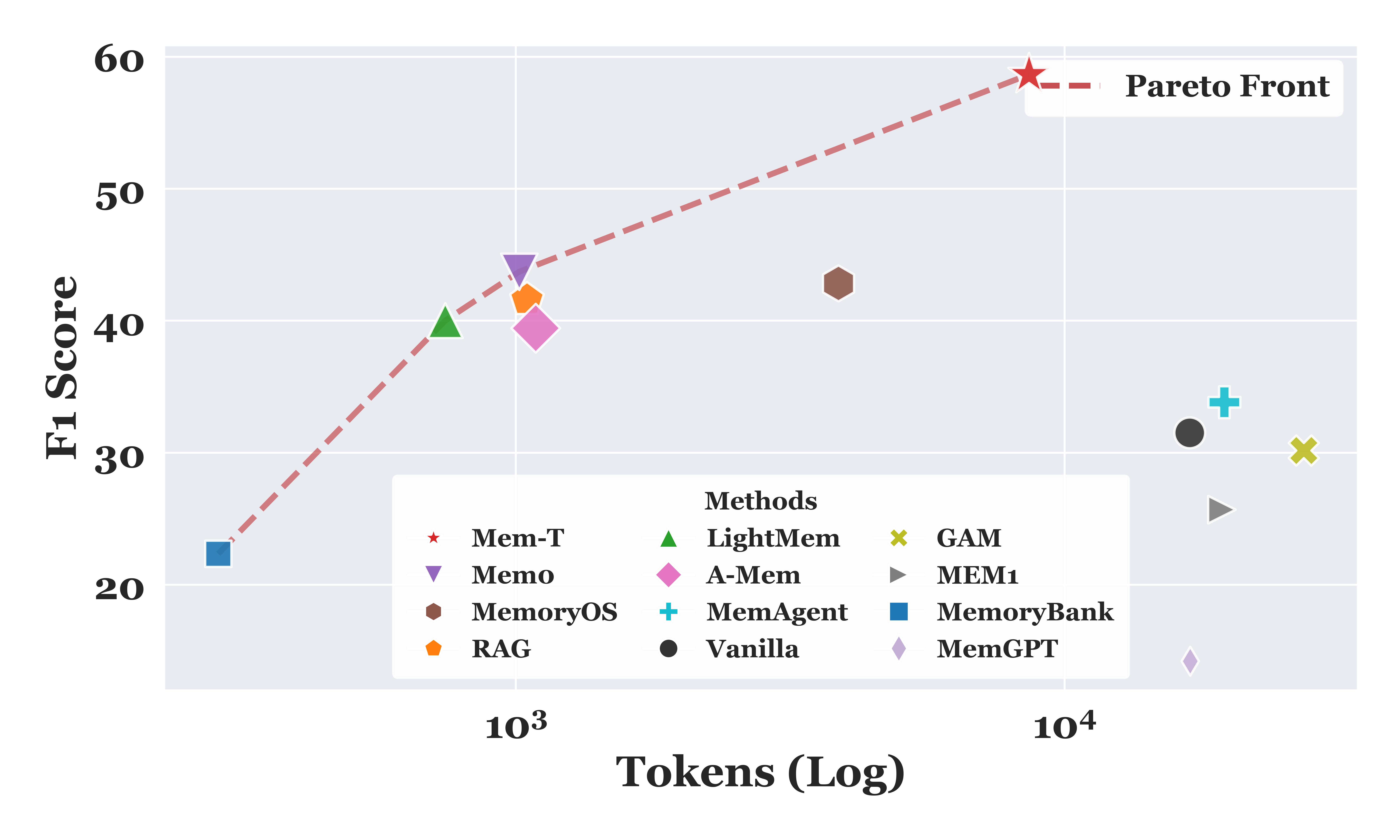}
        \vspace{-1.6em}
        \caption{The comparison of the performance and inference cost on the LoCoMo dataset. Different shapes of the scatter points represent various types of baselines.}
        \label{fig:pareto_LoCoMo}
    \end{minipage}
    \hfill
    \begin{minipage}{0.49\linewidth}
        \centering
        \includegraphics[width=\linewidth]{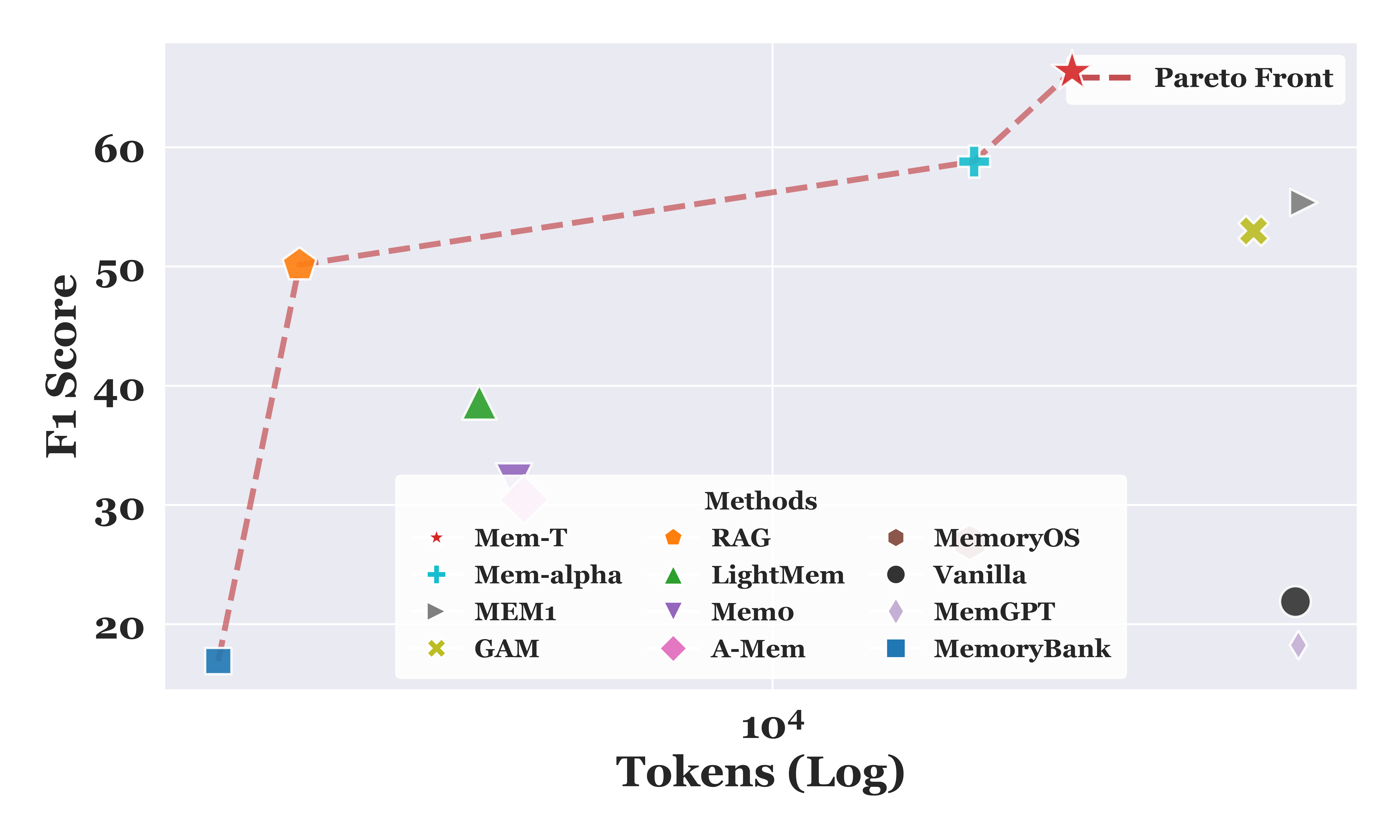}
        \vspace{-1.6em}
        \caption{The comparison of the performance and inference cost on the HotpotQA dataset. Different shapes of the scatter points represent various types of baselines.}
        \label{fig:pareto_hotpotqa}
    \end{minipage}
    \vspace{-0.8em}
\end{figure*}

\paragraph{Token-economical.}
As illustrated in \Cref{fig:pareto_LoCoMo} and \Cref{fig:pareto_hotpotqa}, \ourmethod demonstrates superior cost-effectiveness, lying on the Pareto front for both the LoCoMo and HotpotQA datasets. Compared to GAM, \ourmethod not only achieves a $5.17\sim 28.48$ improvement in F1 Score but also reduces the inference overhead by $19.94\% \sim 24.45\%$ per query.

\subsection{Framework Analysis}

\begin{wraptable}{r}{0.6\linewidth}
    \vspace{-0.8em}
    \centering
    \caption{Ablation study on the LoCoMo dataset. The evaluation metric is set as F1 for all entries.}
    \label{tab:LoCoMo_ablation}
    \resizebox{\linewidth}{!}{%
    \begin{tabular}{l|ccccc}
    \toprule
    \textbf{Method} & \textbf{Single} & \textbf{Multi} & \textbf{Temporal} & \textbf{Open} & \textbf{Overall}\\
    \midrule
    Vanilla \ourmethod & 63.75 & 45.09 & 65.13 & 32.97 & 58.65\\
    \midrule
    \multicolumn{6}{l}{\textit{Ablation of Memory Modules}}\\
    \textit{w/o} $\mathcal{M}_\text{work}$ & 63.24 & 43.42 & 63.38 & 30.90& 57.59\\
    \textit{w/o} $\mathcal{M}_\text{fact}$ & 60.80 & 40.10 & 64.23 & 22.39& 55.25\\
    \textit{w/o} $\mathcal{M}_\text{exp}$  & 61.94 & 43.96 & 62.64 & 27.42& 56.60\\
    \textit{w/o} $\mathcal{M}_\text{raw}$  & 62.19 & 42.84 & 62.41 & 29.38& 56.61\\
    \midrule
    \multicolumn{6}{l}{\textit{Ablation of \ourtrain}}\\
    \textit{w/o} Retr. Opt.& 57.91& 43.85 & 56.69 & 30.73& 53.37 \\
    \textit{w/o} Cons. Opt. & 61.41 & 41.15 & 61.17 & 25.24 & 55.36\\
    \textit{w/o} $A_\text{intra}$ & 62.08 & 43.08 & 63.52 & 31.53 & 56.95\\
    \textit{w/o} $A_\text{inter}$ & 58.33 & 43.52 & 59.58 & 30.59 & 54.09\\
    \bottomrule
    \end{tabular}%
    }
    \vspace{-0.1em}
\end{wraptable}

\paragraph{Ablation Study} 
We conduct an ablation study on the hierarchical memory architecture and the \ourtrain training paradigm, with results presented in \Cref{tab:LoCoMo_ablation}: \textbf{(1) \textit{w/o} Memory Modules}, which individually removes the working ($\mathcal{M}_{\text{work}}$), factual ($\mathcal{M}_{\text{fact}}$), experiential ($\mathcal{M}_{\text{exp}}$), and raw ($\mathcal{M}_{\text{raw}}$) memory stores. On LoCoMo, which emphasizes information extraction in long-horizon dialogues, factual memory proves to be the most critical component, leading to a substantial performance decline of $3.40$. \textbf{(2) \textit{w/o} Optimization Strategies}, where we replace the \ourtrain-optimized policies with the base model during the memory retrieval (\textit{w/o} Retr. Opt.) and construction (\textit{w/o} Cons. Opt.) phases. Eliminating the retrieval optimization leads to the most significant performance decline of $5.28$, while removing the construction optimization causes a $3.29$ drop. These marked degradations verify that both stages of \ourtrain are crucial.
\textbf{(3) \textit{w/o} Advantage Terms}, which ablates the intra-tree ($A_{\text{intra}}$) or inter-tree ($A_{\text{inter}}$) advantage. Removing $A_{\text{inter}}$ causes a larger performance drop ($4.56\downarrow$) than removing $A_{\text{intra}}$ ($1.70\downarrow$), indicating that cross-tree advantage estimation is critical for stable RL training, while combining both signals yields the best performance.  

\begin{figure} \centering \includegraphics[width=0.8\linewidth]{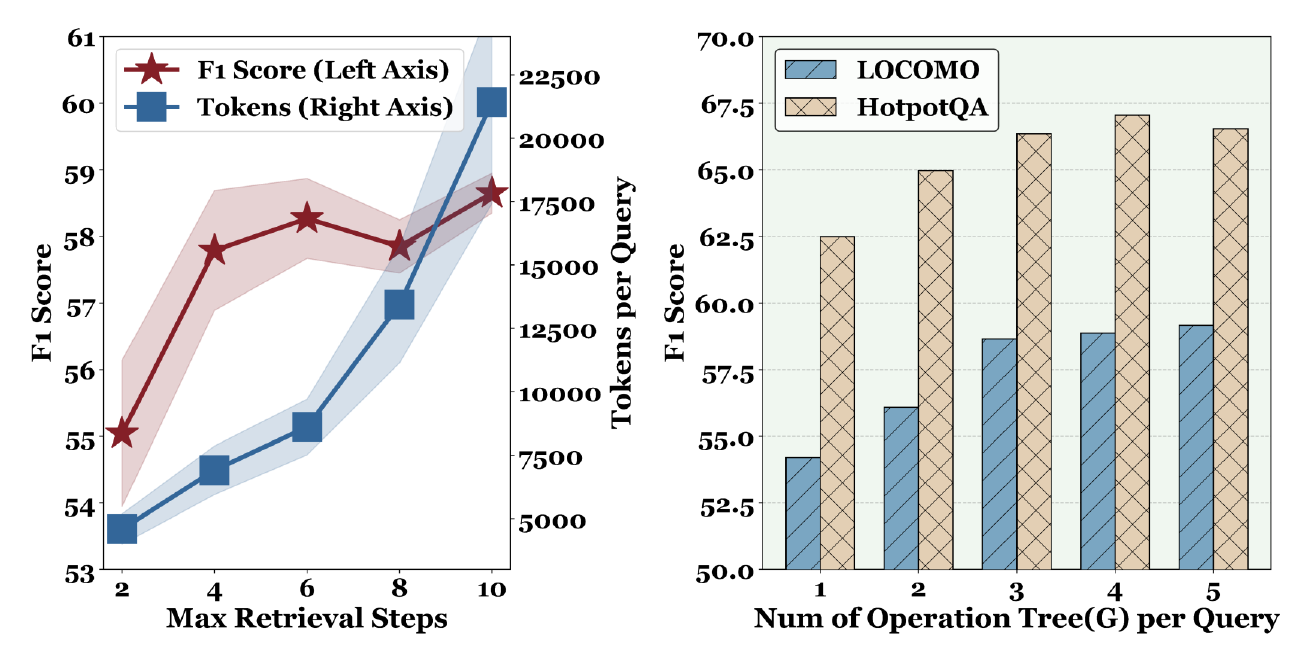} \caption{(\textbf{Left}) Parameter sensitivity analysis on the max inference retrieval steps on the LoCoMo; (\textbf{Right}) Parameter sensitivity analysis on the number of operation trees per query($G$) when training with \ourtrain on the LoCoMo and HotpotQA dataset.} \label{fig:sensi_main} \end{figure}

\paragraph{Sensitivity Analysis}

We analyze the sensitivity of \ourmethod to three core parameters. The results are presented in \Cref{fig:sensi_main} and \Cref{fig:sensi_expansion}.
\textbf{For the maximum retrieval steps}, we observe a substantial performance improvement as the steps increase from $2$ to $6$, where the F1 score increases from $53.45 \rightarrow 58.65$. However, further extending the steps from $6$ to $10$ yields only marginal gains ($<0.5\%$) while linearly inflating the token consumption per query from $\sim 9k$ to $\sim 21k$. 
\textbf{For the number of operation trees $G$}, increasing $G$ from 1 to 3 yields substantial gains, boosting the F1 score on LoCoMo from $54.20$ to $58.65$ and on HotpotQA from $62.49$ to $66.54$. However, further increasing $G$ to 5 results in diminishing returns, offering a marginal average improvement of only $0.35$ while disproportionately inflating the computational cost by approximately $67\%$. 
Thus, we set the maximum retrieval steps to $6$ and $G=3$ to balance efficiency and overhead. More analysis is in \Cref{app:sensi}.

\subsection{Case Study}
\vspace{-0.1em}
\begin{figure}[!t]
    \centering
    \includegraphics[width=\linewidth]{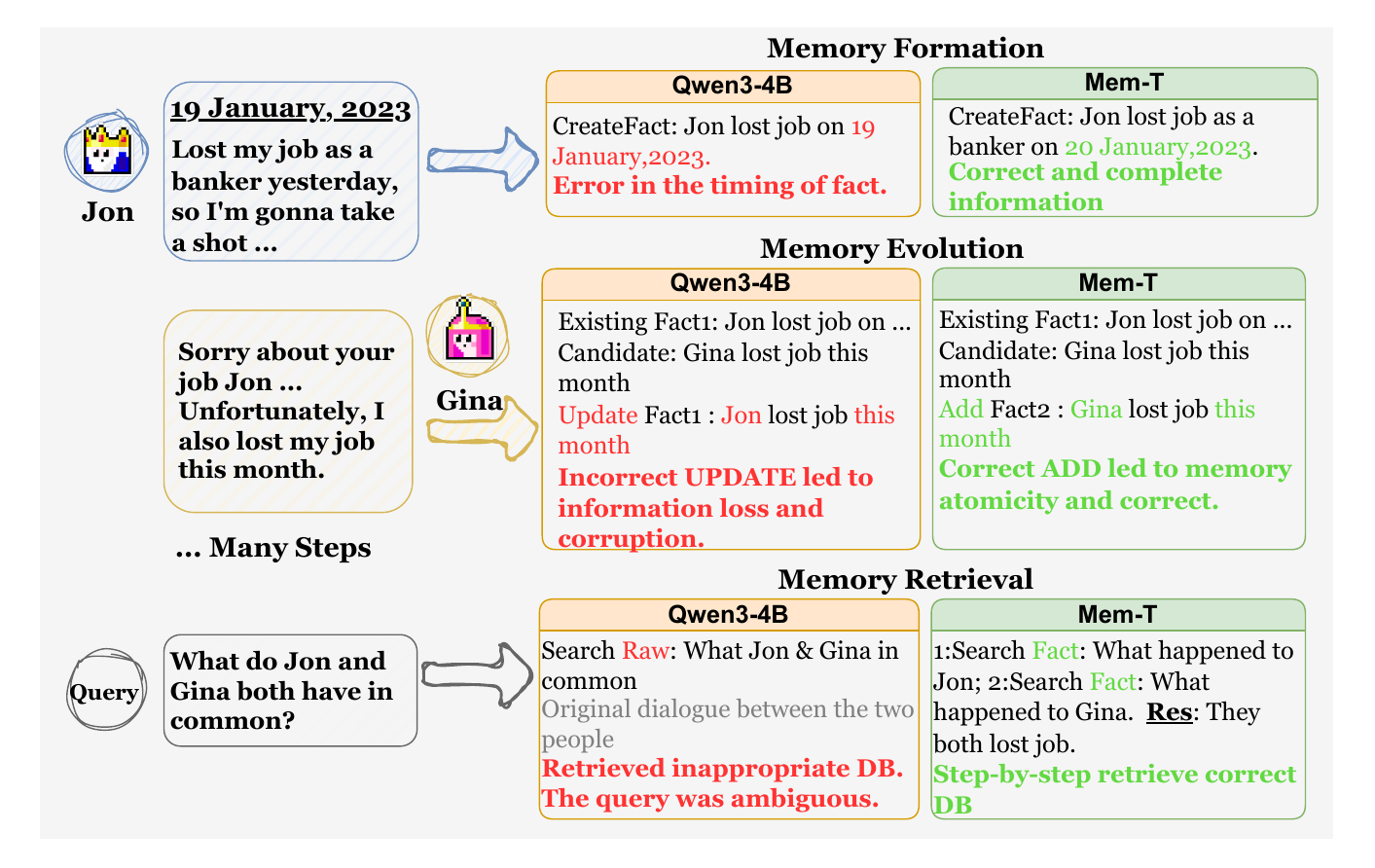}
    \caption{Case Study comparing \ourmethod against baseline.}
    \label{fig:casestudy}
    \vspace{-0.5em}
\end{figure}

We present a case study comparing the memory processing trajectories of \ourmethod against the Qwen3-4B baseline in \Cref{fig:casestudy} to demonstrate the enhanced capabilities acquired through our training paradigm. 

As illustrated, the baseline exhibits severe limitations across the entire memory lifecycle. In the \textbf{formation} phase, it lacks an accurate information extraction capability, failing to resolve relative timestamps (e.g., ``yesterday'') into specific dates. During \textbf{evolution}, it fails to distinguish between \texttt{Update} and \texttt{Add} operations, erroneously overwriting existing entity records with unrelated new memory. Finally, its \textbf{retrieval} mechanism is limited to ambiguous raw queries, lacking the logical depth to handle multi-step reasoning.

In contrast, \ourmethod demonstrates superior capabilities in three aspects:\textbf{\ding{182}Accurate Information Extraction:} It accurately processes raw information (e.g., converting ``yesterday'' to a correct specific date), ensuring initial memory entries are temporally grounded and factually complete; \textbf{\ding{183} Rational Memory Evolution:} It exhibits a deep understanding of the usage criteria for memory evolution tools. By explicitly distinguishing between state updates and new knowledge acquisition, it preserves memory atomicity and prevents key information forgetting.
\textbf{\ding{184} Multi-step Retrieval:} Instead of vague searches, it autonomously decomposes complex queries into sub-questions and retrieves from a suitable store. This step-by-step memory lookups synthesize the answer from distinct memory entries.

\section{Conclusion}
In this paper, we introduce \ourmethod, a comprehensive hierarchical memory framework, and \ourtrain, a novel RL paradigm for memory agents. By decomposing sparse terminal rewards into dense, step-wise supervision via memory operation trees, \ourtrain enables the joint optimization of memory construction and retrieval policies. The extensive experiments demonstrate that \ourmethod not only achieves state-of-the-art performance across in-domain and out-of-domain benchmarks but also realizes a superior Pareto efficiency between task accuracy and inference overhead. We believe \ourmethod represents a shift from heuristic-based storage to fully learnable, attribution-centric memory systems, paving the way for the development of self-evolving agents capable of lifelong learning.

\bibliography{references}

\appendix

\section{Supplementary Experimental Setup}
\subsection{Dataset Description}
\label{app:dataset}
\paragraph{LoCoMo(\citep{maharana2024evaluating})} is a benchmark of very long-term conversational dialogues designed to evaluate long-range memory and reasoning capabilities in agent systems. The dataset consists of 10 extended conversations, each spanning dozens of sessions and hundreds of dialogue turns, with an average of around 600 turns and roughly 16K tokens per conversation. Questions in the LoCoMo QA evaluation are annotated with answer locations and categorized into types such as single-hop, multi-hop, open-domain, temporal reasoning, and adversarial, targeting different memory and inference challenges. In our experiments on LoCoMo QA, we follow standard practice in related work and do not use adversarial question data, which aligns with previous evaluations~\citep{Chhikara2025mem0, xuAMEMAgenticMemory2025}.

\paragraph{Hotpotqa(\citep{yang2018hotpotqadatasetdiverseexplainable})} is a widely-used multi-hop reasoning benchmark that requires models to aggregate information across multiple supporting documents to reach an answer. To evaluate performance in long-context scenarios, we follow the synthesis methodology proposed in recent work~\citep{yu2025memagent,yanGeneralAgenticMemory2025}, where the golden paragraphs containing the necessary evidence are embedded within a haystack of distractor content. In our experiments, we specifically utilize the 56K-token(eval\_400) version of this synthetic HotpotQA dataset. This setup effectively transforms the reasoning task into a long-range retrieval and inference challenge, testing the agent's ability to filter out extensive irrelevant information while maintaining the precision required for multi-step logical reasoning.

\paragraph{LoneMemEval(\citep{DBLP:conf/iclr/WuWYZCY25})} is specifically designed to evaluate the long-term interactive memory capabilities of LLM-driven chat assistants, addressing the underexplored challenge of sustained memory performance in prolonged user-AI interactions. It comprehensively assesses five core memory abilities, information extraction, multi-session reasoning, temporal reasoning, knowledge updates, and abstention, through 500 manually curated questions embedded in freely scalable user-assistant chat histories, with two standard configurations: LONGMEMEVALS ($115k$ tokens per question) and LONGMEMEVALM ($\sim1.5$ million tokens across 500 sessions). Following previous works~\citep{DBLP:journals/corr/abs-2509-25911,Rasmussen2025Zep,fang2025lightmem}, we use the LONGMEMEVALS dataset.

\paragraph{NarrativeQA(\citep{kočiský2017narrativeqareadingcomprehensionchallenge})} is a large-scale reading comprehension benchmark that assesses models’ ability to understand and reason over long narrative text, such as books and movie scripts. The full NarrativeQA dataset contains on the order of tens of thousands of human-written question–answer pairs associated with over a thousand story documents, where questions require synthesis across global document structure rather than shallow pattern matching. Questions are constructed based on human-generated abstractive summaries, encouraging deep narrative understanding and integrative reasoning beyond local context overlaps. Following \citep{hu2026memorymattersmoreeventcentric}, we randomly sampled 10 long documents from the NarrativeQA corpus and used their associated 298 QA pairs to measure performance on long-range narrative question answering. 

\subsection{Implementation Details}
\label{app:imple}
\paragraph{\ourtrain for Memory Retrieval}

\noindent\textit{Training Implementation Details.} 
We utilize the Ray distributed framework combined with vLLM as the inference backend, employing XFormers to optimize attention mechanisms. The model is trained with a global batch size of 32. We adopt a peak learning rate of $5 \times 10^{-6}$ with a warmup ratio of $0.285$. To ensure training stability and prevent reward hacking, we set the KL divergence coefficient to $0.001$. 

\noindent\textit{Context and Efficiency.} 
To support extensive memory retrieval operations, we configure the system with an extended context window, allowing for a maximum prompt length of 40,960 tokens and a maximum observation history of 20,480 tokens. For computational efficiency, we employ Fully Sharded Data Parallel (FSDP) with parameter, gradient, and optimizer offloading, performing all computations in \texttt{bfloat16} precision.

\paragraph{\ourtrain for Memory Construction}

\noindent\textit{Training Configuration.} 
The training is conducted on a single node equipped with 8 GPUs, utilizing the LLaMA-Factory framework. To maximize computational efficiency and handle the memory footprint of full-parameter updates, we employ DeepSpeed ZeRO-3 combined with Flash Attention 2. The maximum sequence length is truncated to $6,144$ tokens.

\noindent\textit{Hyperparameters.} 
The global batch size is set to $32$ (calculated with a per-device batch size of $2$ and $2$ gradient accumulation steps). We optimize the model for $200$ steps using a cosine learning rate scheduler, with a peak learning rate of $5 \times 10^{-6}$ and a warmup ratio of $0.1$. The training uses \texttt{bfloat16} precision, and $10\%$ of the dataset is reserved for validation to monitor convergence.

\section{Supplementary Experiment}
\subsection{Generalization Experiments Across Other LLMs}
\label{app:qwen8B}
\begin{table*}[!ht]
    \centering
    \caption{Performance comparison on the LoCoMo benchmark, with F1 and BLEU-1 as the evaluation metrics. $^{\ddagger}$: As Memory-R1 is not open-source, we faithfully report the results provided in their original paper.}
    \label{tab:LoCoMo_results_8B}
    \resizebox{\textwidth}{!}{%
    \begin{tabular}{llcccccccccc}
        \toprule
        \multirow{2}{*}{\textbf{Method}} & \multirow{2}{*}{\textbf{LLM}} & \multicolumn{2}{c}{\textbf{Single-Hop}} & \multicolumn{2}{c}{\textbf{Multi-Hop}} & \multicolumn{2}{c}{\textbf{Temporal}} & \multicolumn{2}{c}{\textbf{Open Domain}} & \multicolumn{2}{c}{\textbf{Overall}} \\
        \cmidrule(lr){3-4} \cmidrule(lr){5-6} \cmidrule(lr){7-8} \cmidrule(lr){9-10} \cmidrule(lr){11-12}
         & & F1$\uparrow$ & B1$\uparrow$ & F1$\uparrow$ & B1$\uparrow$ & F1$\uparrow$ & B1$\uparrow$ & F1$\uparrow$ & B1$\uparrow$ & F1$\uparrow$ & B1$\uparrow$ \\
        \midrule
        \multicolumn{12}{l}{\textit{Training-free Methods}} \\
        RAG  & Qwen3-8B & 49.62 & 43.98 & 23.64 & 17.82 & 37.93 & 33.80 & 21.39 & 16.33 & 40.77 & 35.43 \\
        MemGPT & Qwen3-8B & 16.23 & 13.08 & 18.13 & 13.72 & 15.87 & 11.39 & 14.18 & 10.66 & 16.38 & 12.71 \\
        MemoryBank & Qwen3-8B & 26.50 & 19.48 & 26.52 & 18.93 & 15.49 & 11.36 & 15.92 & 12.09 & 23.66 & 17.31 \\
        Mem0   & Qwen3-8B & 45.92 & 39.93 & 27.80 & 19.97 & 43.64 & 33.82 & 18.37 & 13.84 & 40.41 & 33.42 \\
        MemoryOS  & Qwen3-8B & 48.77 & 43.47 & 29.19 & 24.87 & 42.98 & 35.27 & 18.50 & 15.09 & 42.12 & 36.65 \\
        LightMem  & Qwen3-8B & 49.89 & 44.48 & 33.98 & 27.60 & 44.53 & 39.65 & 19.37 & 14.05 & 43.98 & 38.51 \\
        A-Mem & Qwen3-8B & 47.75 & 41.36 & 32.35 & 24.82 & 36.80 & 30.71 & 18.62 & 14.98 & 40.92 & 34.56 \\
        GAM   & Qwen3-8B & 46.62 & 40.15 & 32.18 & 24.96 & 46.42 & 39.71 & 13.56& 10.32 & 41.84 & 35.39 \\
        \midrule
        \multicolumn{12}{l}{\textit{Trained Methods}} \\
        MEM1  & MEM1-7B & 27.48 & 22.10 & 18.98 & 15.56 & 30.52 & 23.48 & 14.21 & 11.43 & 25.68 & 20.50 \\
        MemAgent & MemAgent-14B & 35.86 & 29.64 & 27.86 & 22.72 & 37.93 & 31.85 & 20.31 & 16.47 & 33.82 & 27.97 \\
        Memory-R1-PPO$^{\ddagger}$  & Mem-R1-8B & 32.52 & 24.47 & 26.86 & 23.47 & 41.57 & 26.11 & \underline{45.30} & \underline{39.18} & 34.08 & 25.54 \\
        Memory-R1-GRPO$^{\ddagger}$  & Mem-R1-8B & 35.73 & 27.70 & 35.65 & 30.77 & 49.86 & 38.27 & \textbf{47.42} & \textbf{41.24} & 39.25 & 31.21 \\
        \midrule
        \multicolumn{12}{l}{\textit{Our Method}} \\
        w/o training & Qwen3-8B & \underline{55.89} & \underline{51.14} & \underline{38.13} & 30.33 & \underline{53.30} & \underline{47.02} & 23.55 & 20.18 & \underline{50.08} & \underline{44.55} \\
        with \ourtrain & Qwen3-8B & \textbf{63.65} & \textbf{57.97} & \textbf{42.38} & \textbf{34.72} & \textbf{66.85} & \textbf{62.29} & 34.33 & 31.47 & \textbf{58.53} & \textbf{52.89} \\
        \bottomrule
    \end{tabular}%
    }
\end{table*}

\Cref{tab:LoCoMo_results_8B} demonstrates the generalization capabilities of \ourmethod when applied to the Qwen3-8B model. The results indicate that our approach significantly outperforms all existing baselines across most metrics in the LoCoMo benchmark. Notably, even our training-free variant achieves an Overall F1 score of 50.08, surpassing previously established trained models such as Memory-R1 and MemAgent. When combined with our specific training, the performance further improves to 58.53 F1, particularly excelling in Single-Hop and Temporal reasoning tasks, thereby confirming the robust transferability and effectiveness of our framework across different LLM backbones.

\subsection{Sensitivity Analysis}
\label{app:sensi}
\begin{figure}[!ht]
    \centering
    \includegraphics[width=0.5\linewidth]{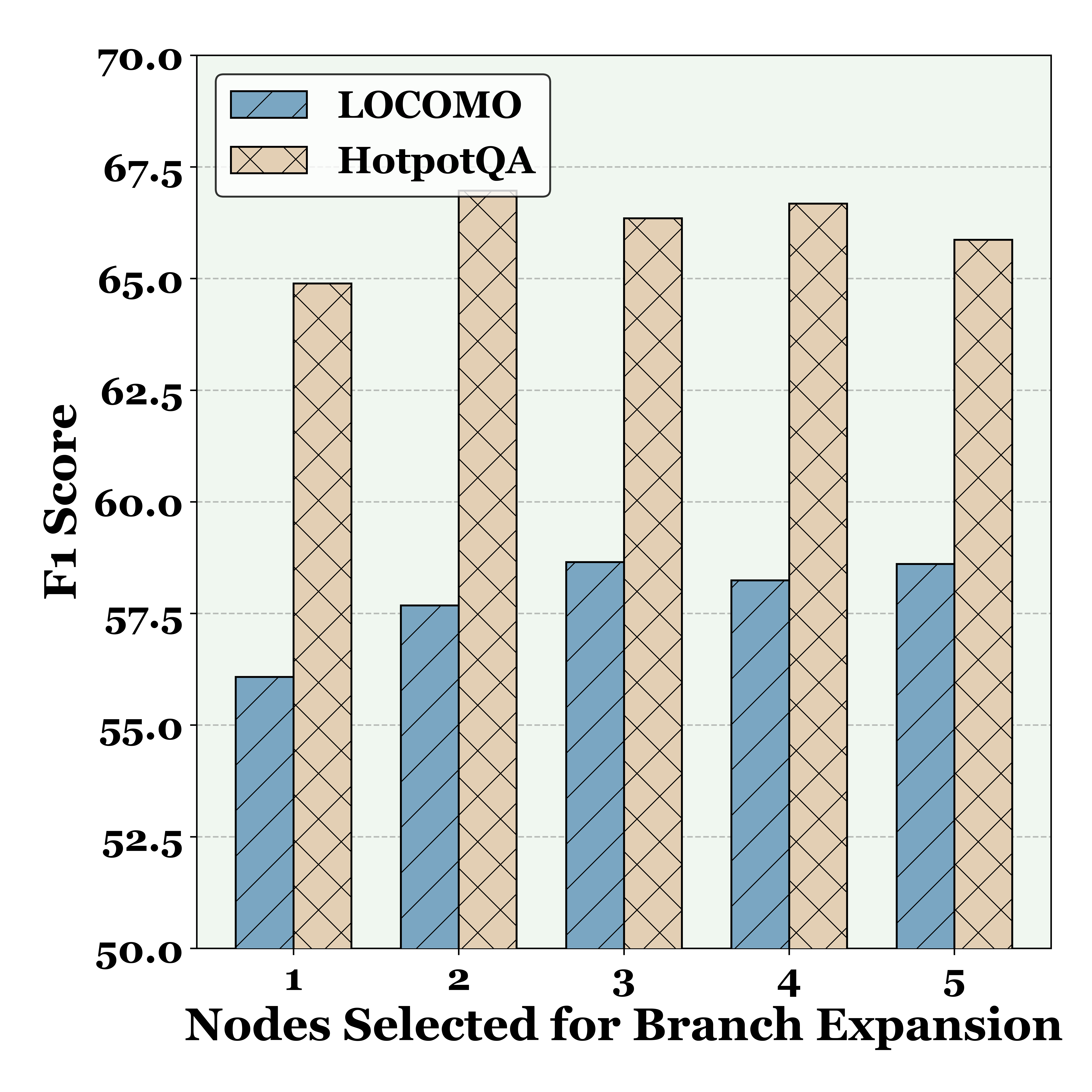}
    \caption{Parameter sensitivity analysis on the number of nodes selected for branch expansion when training with \ourtrain on the LoCoMo and HotpotQA dataset.}
    \label{fig:sensi_expansion}
\end{figure}
\textbf{Regarding the number of nodes selected for branch expansion}, as shown in \Cref{fig:sensi_expansion}, we observe that increasing the number of nodes selected for branch expansion from 1 to 3 leads to significant performance improvements, with the F1 score rising from 56.08 to 58.65 on LoCoMo and from 64.89 to 66.35 on HotpotQA. However, further increasing the expansion breadth beyond 3 nodes yields diminishing returns; for instance, at a node count of 5, the F1 scores for both datasets plateau or even slightly decrease. Given that a larger number of expansion nodes significantly increases the search space and computational latency, we select 3 as the optimal number of nodes for branch expansion to achieve the best trade-off between reasoning accuracy and inference efficiency.

\subsection{Training Curves}
\begin{figure}[!ht]
    \centering
    \includegraphics[width=0.5\linewidth]{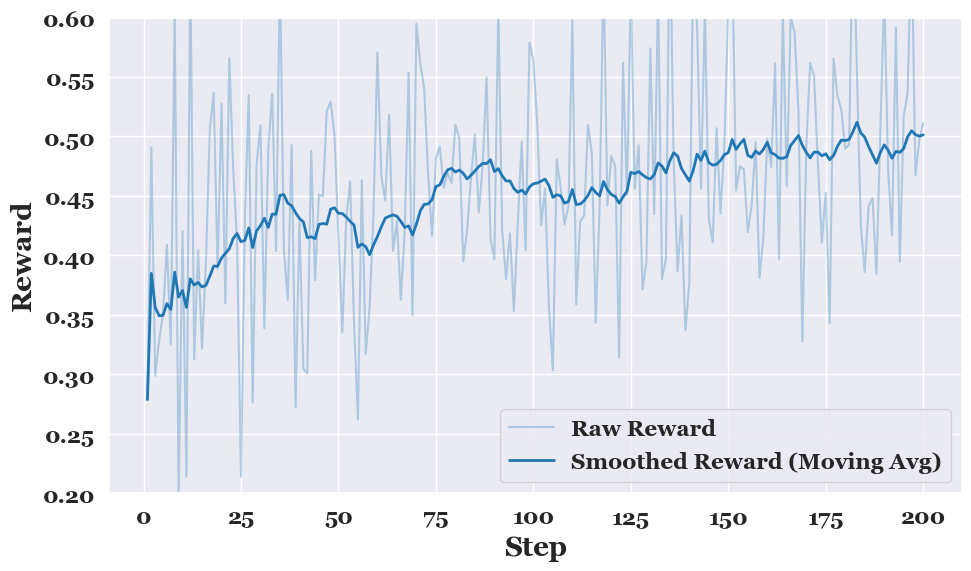}
    \caption{Reward curves of memory retrieval training under \ourtrain.}
    \label{fig:reward}
\end{figure}
\textbf{Regarding the memory retrieval training stage}, \Cref{fig:reward} illustrates the evolution of rewards under the \ourtrain framework. The smoothed reward curve exhibits a consistent upward trend, climbing from an initial value of approximately $0.30$ to over $0.50$ by the 200th step. Although the raw rewards show significant variance, typical of reinforcement learning in complex reasoning tasks, the steady improvement in the moving average confirms that the agent effectively learns to optimize its retrieval strategies to maximize task-specific gains.

\begin{figure}[!ht]
    \centering
    \includegraphics[width=0.5\linewidth]{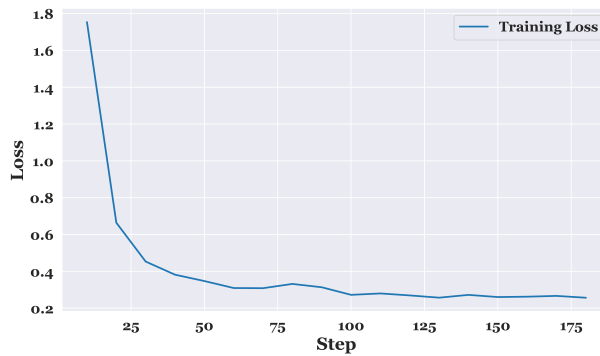}
    \caption{Loss curves of memory construction training under \ourtrain.}
    \label{fig:loss}
\end{figure}
\textbf{For the memory construction training stage}, \Cref{fig:loss} presents the training loss over 180 steps. The curve shows a sharp initial descent, with the loss dropping rapidly from $1.8$ to below $0.4$ within the first 50 steps, indicating efficient convergence. In the subsequent phase, the loss stabilizes and fluctuates marginally around $0.25$, suggesting that the model has successfully captured the underlying patterns for memory synthesis and state updates. The overall stability of the loss curve demonstrates the robustness of the memory construction process under our policy optimization framework.
\section{Prompts of \ourmethod}
\subsection{Memory Formation}
\begin{tcolorbox}[title={CreateFactTool}, sharp corners, breakable, 
      colframe=Periwinkle, colback=white, 
        boxrule=3pt, boxsep=0.5pt, enhanced, 
        shadow={3pt}{-3pt}{0pt}{opacity=1}]
        \footnotesize
        {\fontfamily{pcr}\selectfont
\begin{lstlisting}[breaklines=true,showstringspaces=false]
class CreateFactTool(BaseTool):
    def __init__(self):
        super().__init__(
            name="create_fact",
            description=(
                "Extract 'Factual Memory' (Concrete, verifiable statements about WHAT happened).\n"
                "CRITICAL RULES:\n"
                "1. Full Entity Scan: Extract attributes and relationships between all entities mentioned (e.g., specific objects, places, third parties), not just the user.\n"
                "2. Pay special attention to time information, including relative times like 'yesterday' or 'the week before' in the text.\n"
                "Target two specific types of facts:\n"
                "1. User Factual Memory: Verifiable facts about the user's events experienced, identity, preferences, items owned, and specific constraints.\n"
                "2. Environment Factual Memory: Explicit states of the external world, object properties, document knowledge, or tool states, and other entities.\n"
            ),
            parameters={
                "fact": {
                    "type": "string", 
                    "description": "The concise, standalone declarative statement. E.g., 'The user prefers Python for backend tasks' or 'The API endpoint v2 is deprecated'."
                },
                "start_time": {
                    "type": "string", 
                    "description": "The time when the event occurred or the time when the attribute is valid. Use an empty string if it does not exist."
                },
                "end_time": {
                    "type": "string", 
                    "description": "The end time of the event or the expiration time of the attribute. Use an empty string if it does not exist."
                }
            },
            required=["fact", "start_time", "end_time"]
        )
\end{lstlisting}
}
\end{tcolorbox}

\begin{tcolorbox}[title={CreateExperienceTool}, sharp corners, breakable, 
      colframe=Periwinkle, colback=white, 
        boxrule=3pt, boxsep=0.5pt, enhanced, 
        shadow={3pt}{-3pt}{0pt}{opacity=1}]
        \footnotesize
        {\fontfamily{pcr}\selectfont
\begin{lstlisting}[breaklines=true,showstringspaces=false]
class CreateExperienceTool(BaseTool):
    def __init__(self):
        super().__init__(
            name="create_experience",
            description="Extract 'Experiential Memory' (Actionable lessons, patterns, or HOW-TO perform a task)\n"
                "This tool captures lessons learned, reasoning patterns, and executable skills.\n"
                "1. Strategy-based: Reusable heuristics, workflows, or insights derived from reasoning (e.g., 'To solve X, method Y is most efficient').\n"
                "2. Case-based: Key trajectories of Success or Failure that serve as examples (e.g., 'Attempting action A under condition B caused error C').\n"
                "3. Skill-based: Validated code snippets, tool usage protocols, or functions that the agent can execute.\n",
            parameters={
                "experience": {
                    "type": "string", 
                    "description": "The distilled content of the experience. It should be formulated as a rule, a cause-effect relationship, or a guideline for future actions."
                    },
                "start_time": {
                    "type": "string", 
                    "description": "The time when the event occurred or the time when the attribute is valid. Use an empty string if it does not exist."
                    },
                "end_time": {
                    "type": "string", 
                    "description": "The end time of the event or the expiration time of the attribute. Use an empty string if it does not exist."
                    }
            },
            required=["experience","start_time","end_time"]
        )
\end{lstlisting}
}
\end{tcolorbox}

\begin{tcolorbox}[title={UpdatePersonaTool}, sharp corners, breakable, 
      colframe=Periwinkle, colback=white, 
        boxrule=3pt, boxsep=0.5pt, enhanced, 
        shadow={3pt}{-3pt}{0pt}{opacity=1}]
        \footnotesize
        {\fontfamily{pcr}\selectfont
\begin{lstlisting}[breaklines=true,showstringspaces=false]
class UpdatePersonaTool(BaseTool):
    def __init__(self):
        super().__init__(
            name="update_persona",
            description="If there is new and important information about the person, such as hobbies, participated projects or significant events, update(add or modify) the full character profile for that person.",
            parameters={
                "name": {
                    "type": "string", 
                    "description": "Name of the person. Or 'User' if the user does not have a specified name."
                    },
                "profile": {
                    "type": "string", 
                    "description": "The full, concise and updated persona text."
                    }
            },
            required=["name", "profile"]
        )
\end{lstlisting}
}
\end{tcolorbox}

\begin{tcolorbox}[title={UpdateSummaryTool}, sharp corners, breakable, 
      colframe=Periwinkle, colback=white, 
        boxrule=3pt, boxsep=0.5pt, enhanced, 
        shadow={3pt}{-3pt}{0pt}{opacity=1}]
        \footnotesize
        {\fontfamily{pcr}\selectfont
\begin{lstlisting}[breaklines=true,showstringspaces=false]
class UpdateSummaryTool(BaseTool):
    def __init__(self):
        super().__init__(
            name="update_summary",
            description="If there is new information based on the current conversation, update the runtime summary of the sessions.",
            parameters={
                "content": {
                    "type": "string",
                    "description": "The concise, complete and updated summary text."
                    }
            },
            required=["content"]
        )
\end{lstlisting}
}
\end{tcolorbox}

\subsection{Memory Evolution}
\begin{tcolorbox}[title={AddItemTool}, sharp corners, breakable, 
      colframe=Periwinkle, colback=white, 
        boxrule=3pt, boxsep=0.5pt, enhanced, 
        shadow={3pt}{-3pt}{0pt}{opacity=1}]
        \footnotesize
        {\fontfamily{pcr}\selectfont
\begin{lstlisting}[breaklines=true,showstringspaces=false]
class AddItemTool(BaseTool):
    def __init__(self, vector_db: VectorDBBase, collection_name: str):
        super().__init__(
            name="add_item",
            description="Add a new memory item.",
            parameters={
                "document": {"type": "string", "description": "The content."},
                "turn_time": {"type": "string", "description": "The time of the turn that generated this item."},
                "start_time": {"type": "string", "description": "Start time."},
                "end_time": {"type": "string", "description": "End time."},
            },
            required=["document"]
        )
        self.db = vector_db
        self.collection_name = collection_name
\end{lstlisting}
}
\end{tcolorbox}

\begin{tcolorbox}[title={UpdateItemTool}, sharp corners, breakable, 
      colframe=Periwinkle, colback=white, 
        boxrule=3pt, boxsep=0.5pt, enhanced, 
        shadow={3pt}{-3pt}{0pt}{opacity=1}]
        \footnotesize
        {\fontfamily{pcr}\selectfont
\begin{lstlisting}[breaklines=true,showstringspaces=false]
class UpdateItemTool(BaseTool):
    def __init__(self, vector_db: VectorDBBase, collection_name: str):
        super().__init__(
            name="update_item",
            description="Update an existing memory item.",
            parameters={
                "id": {"type": "string", "description": "The ID of the item to update."},
                "document": {"type": "string", "description": "Enrich the content with more details and update the statistical data or factual frequencies mentioned. Must save the original time information of previously items in the document."},
                "turn_time": {"type": "string", "description": "The time of the turn that generated this update."},
                "start_time": {"type": "string", "description": "New start time."},
                "end_time": {"type": "string", "description": "New end time."},
            },
            required=["id", "document"]
        )
        self.db = vector_db
        self.collection_name = collection_name
\end{lstlisting}
}
\end{tcolorbox}

\begin{tcolorbox}[title={DeleteItemTool}, sharp corners, breakable, 
      colframe=Periwinkle, colback=white, 
        boxrule=3pt, boxsep=0.5pt, enhanced, 
        shadow={3pt}{-3pt}{0pt}{opacity=1}]
        \footnotesize
        {\fontfamily{pcr}\selectfont
\begin{lstlisting}[breaklines=true,showstringspaces=false]
class DeleteItemTool(BaseTool):
    def __init__(self, vector_db: VectorDBBase, collection_name: str):
        super().__init__(
            name="delete_item",
            description="Delete an existing memory item. Use when an item is explicitly negated or wrong.",
            parameters={
                "id": {"type": "string", "description": "The ID to delete."}
            },
            required=["id"]
        )
        self.db = vector_db
        self.collection_name = collection_name
\end{lstlisting}
}
\end{tcolorbox}

\begin{tcolorbox}[title={IgnoreItemTool}, sharp corners, breakable, 
      colframe=Periwinkle, colback=white, 
        boxrule=3pt, boxsep=0.5pt, enhanced, 
        shadow={3pt}{-3pt}{0pt}{opacity=1}]
        \footnotesize
        {\fontfamily{pcr}\selectfont
\begin{lstlisting}[breaklines=true,showstringspaces=false]
class IgnoreItemTool(BaseTool):
    def __init__(self):
        super().__init__(
            name="ignore_item",
            description="Do nothing. If the item is completely redundant in both *semantic meaning* and *time range*.",
            parameters={
                 "reason": {"type": "string", "description": "Reason for ignoring."}
            },
            required=["reason"]
        )
\end{lstlisting}
}
\end{tcolorbox}

\subsection{Memory Retrieval}
\begin{tcolorbox}[title={SearchSummaryTool}, sharp corners, breakable, 
      colframe=Periwinkle, colback=white, 
        boxrule=3pt, boxsep=0.5pt, enhanced, 
        shadow={3pt}{-3pt}{0pt}{opacity=1}]
        \footnotesize
        {\fontfamily{pcr}\selectfont
\begin{lstlisting}[breaklines=true,showstringspaces=false]
class SearchSummaryTool(BaseTool):
    def __init__(self, vector_db: VectorDBBase):
        super().__init__(
            name="search_summary",
            description="Retrieve relevant summaries to quickly understand the context background.",
            parameters={"query": {"type": "string", "description": "Query string."}},
            required=["query"]
        )
        self.db = vector_db
\end{lstlisting}
}
\end{tcolorbox}

\begin{tcolorbox}[title={SearchFactsTool}, sharp corners, breakable, 
      colframe=Periwinkle, colback=white, 
        boxrule=3pt, boxsep=0.5pt, enhanced, 
        shadow={3pt}{-3pt}{0pt}{opacity=1}]
        \footnotesize
        {\fontfamily{pcr}\selectfont
\begin{lstlisting}[breaklines=true,showstringspaces=false]
class SearchFactsTool(BaseTool):
    def __init__(self, vector_db: VectorDBBase, top_k: int):
        super().__init__(
            name="search_facts",
            description = "Retrieve 'Factual Memory' (Concrete, verifiable statements about WHAT happened).\n"
                "Target two specific types of facts:\n"
                "1. User Factual Memory: Verifiable facts about the user's identity, stable preferences, important events, habits, "
                "historical commitments, and specific constraints.\n"
                "2. Environment Factual Memory: Explicit states of the external world, object properties, "
                "document knowledge, or tool states.\n",
            parameters = {"query": {"type": "string",
                                   "description": "A self-contained, semantically rich search query rewritten from the user's intent.\n"
                                                  "Instead of raw questions like 'Does he like it?', use specific declarative queries like "
                                                  "'User preference regarding spicy food' or 'Attributes of Object X'."}},
            required=["query"]
        )
        self.db = vector_db
        self.top_k = top_k
\end{lstlisting}
}
\end{tcolorbox}

\begin{tcolorbox}[title={SearchExperiencesTool}, sharp corners, breakable, 
      colframe=Periwinkle, colback=white, 
        boxrule=3pt, boxsep=0.5pt, enhanced, 
        shadow={3pt}{-3pt}{0pt}{opacity=1}]
        \footnotesize
        {\fontfamily{pcr}\selectfont
\begin{lstlisting}[breaklines=true,showstringspaces=false]
class SearchExperiencesTool(BaseTool):
    def __init__(self, vector_db: VectorDBBase, top_k: int):
        super().__init__(
            name="search_experiences",
            description=(
                "Extract 'Experiential Memory' (Actionable lessons, patterns, or HOW-TO perform a task)\n"
                "This tool captures lessons learned, reasoning patterns, and executable skills.:\n"
                "1. Strategy-based: Reusable heuristics, workflows, or insights derived from reasoning (e.g., 'To solve X, method Y is most efficient').\n"
                "2. Case-based: Key trajectories of Success or Failure that serve as examples (e.g., 'Attempting action A under condition B caused error C').\n"
                "3. Skill-based: Validated code snippets, tool usage protocols, or functions that the agent can execute.\n"
                "Avoid recording raw dialogue history; focus on the distilled 'Lesson' or 'Rule'."
            ),
            parameters={
                "query": {
                    "type": "string",
                    "description": (
                        "A self-contained, semantically rich search query rewritten from the user's intent. \n"
                        "Formulate problem-solving queries like 'Standard workflow for analyzing finance reports' "
                        "or 'How to handle TimeoutError in API calls'."
                    )
                }
            },
            required=["query"]
        )
        self.db = vector_db
        self.top_k = top_k
\end{lstlisting}
}
\end{tcolorbox}

\begin{tcolorbox}[title={SearchPersonasTool}, sharp corners, breakable, 
      colframe=Periwinkle, colback=white, 
        boxrule=3pt, boxsep=0.5pt, enhanced, 
        shadow={3pt}{-3pt}{0pt}{opacity=1}]
        \footnotesize
        {\fontfamily{pcr}\selectfont
\begin{lstlisting}[breaklines=true,showstringspaces=false]
class SearchPersonasTool(BaseTool):
    def __init__(self, vector_db: VectorDBBase):
        super().__init__(
            name="search_personas",
            description="Retrieve character profiles or insights for specific individuals.",
            parameters={
                "name": {"type": "string", "description": "Name of the target individual for exact lookup."},
                "query": {"type": "string", "description": "Query string to find personas by traits; ignored if 'name' is provided."}
            },
            required=["query"]
        )
        self.db = vector_db
\end{lstlisting}
}
\end{tcolorbox}

\begin{tcolorbox}[title={SearchTurnsTool}, sharp corners, breakable, 
      colframe=Periwinkle, colback=white, 
        boxrule=3pt, boxsep=0.5pt, enhanced, 
        shadow={3pt}{-3pt}{0pt}{opacity=1}]
        \footnotesize
        {\fontfamily{pcr}\selectfont
\begin{lstlisting}[breaklines=true,showstringspaces=false]
class SearchTurnsTool(BaseTool):
    def __init__(self, vector_db: VectorDBBase, top_k: int):
        super().__init__(
            name="search_turns",
            description="Retrieve specific raw dialogue history (Raw Turns). \n"
                        "Use this tool for questions about specific past conversations, verifying exact quotes, or checking 'what was' in detail. \n"
                        "Raw turns provide the most authentic context that summaries or facts might miss.",
            parameters={
                "query": {"type": "string", "description": "Keywords or specific quotes."},
                "top_k": {"type": "integer", "description": "The number of turns to retrieve. Default is 5."}
                },
            required=["query"]
        )
        self.db = vector_db
\end{lstlisting}
}
\end{tcolorbox}

\begin{tcolorbox}[title={FinishTool}, sharp corners, breakable, 
      colframe=Periwinkle, colback=white, 
        boxrule=3pt, boxsep=0.5pt, enhanced, 
        shadow={3pt}{-3pt}{0pt}{opacity=1}]
        \footnotesize
        {\fontfamily{pcr}\selectfont
\begin{lstlisting}[breaklines=true,showstringspaces=false]
class FinishTool(BaseTool):
    def __init__(self, benchmark_name: str = "locomo", category: str = ""):
        self.benchmark_name = benchmark_name
        self.category = str(category) if category else ""
        
        description = "Call this when you are confident that you can give the correct final answer. Or you should continue to retrieve more information."
        
        super().__init__(
            name="finish",
            description=description,
            parameters={"answer": {"type": "string", "description": "The concise answer following the Final Result Format."}},
            required=["answer"]
        )
\end{lstlisting}
}
\end{tcolorbox}

\end{document}

%% file: notations.tex
\usepackage{xspace}

\theoremstyle{plain}

\newtheorem*{proposition*}{Proposition}

\theoremstyle{definition}

\theoremstyle{definition}

\def\eqref#1{equation~\ref{#1}}

